%% file: 1_wacv_Paper.tex
\documentclass[10pt,twocolumn,letterpaper]{article}

\usepackage[english]{babel}
\usepackage[shortlabels]{enumitem}
\usepackage[table,xcdraw]{xcolor}
\usepackage{aux_files/slashbox}
\input{aux_files/macros}
\usepackage[boxruled, vlined, linesnumbered]{algorithm2e}
\usepackage{lmodern,babel,adjustbox,booktabs,multirow}
\usepackage{float}
\SetAlFnt{\small}
\SetAlCapFnt{\small}
\SetAlCapNameFnt{\small}
\usepackage{algorithmic}
\algsetup{linenosize=\tiny}
\usepackage{subcaption}
\usepackage{bbm, bm}
\usepackage{makecell}

\usepackage{placeins}

\usepackage{aux_files/wacv}
\usepackage{graphicx}
\usepackage{amsmath}
\usepackage{amssymb}
\usepackage{booktabs}
\usepackage{sidecap}

\usepackage[pagebackref,breaklinks,colorlinks]{hyperref}
\usepackage{xurl} 

\usepackage[capitalize]{cleveref}
\crefname{section}{Sec.}{Secs.}
\Crefname{section}{Section}{Sections}
\Crefname{table}{Table}{Tables}
\crefname{table}{Tab.}{Tabs.}

\begin{document}

\title{A Realistic Protocol for Evaluation of Weakly Supervised Object Localization}

\author{Shakeeb~Murtaza,
 ~Soufiane~Belharbi,
  ~Marco~Pedersoli,
  ~Eric~Granger
  \\
  LIVIA, ILLS, Dept. of Systems Engineering, ETS Montreal, Canada\\
  {\tt\footnotesize \textcolor{black}{\{shakeeb.murtaza.1, soufiane.belharbi.1\}@ens.etsmtl.ca}, \{marco.pedersoli, eric.granger\}@etsmtl.ca }
}
\maketitle

\AddToShipoutPicture*{%
  \AtPageUpperLeft{%
    \put(\LenToUnit{0.5\paperwidth},\LenToUnit{-0.2in}){
      \makebox(0,0)[c]{%
        \begin{minipage}{0.86\paperwidth}
          \centering\small
          This paper has been published in the proceedings of WACV 2025.
          The full text is available at
          \href{https://openaccess.thecvf.com/content/WACV2025/papers/Murtaza_A_Realistic_Protocol_for_Evaluation_of_Weakly_Supervised_Object_Localization_WACV_2025_paper.pdf}{CVF}.
        \end{minipage}%
      }%
    }%
  }%
}%

\begin{abstract}
\input{sections/0_abstract_content}
\end{abstract}

\input{sections/1_main_content}

{\small
\bibliographystyle{aux_files/ieee_fullname}
\bibliography{aux_files/main}
}

\clearpage
\appendix

\numberwithin{equation}{section}
\numberwithin{figure}{section}
\numberwithin{table}{section}

\clearpage

\twocolumn[
\begin{center}
{\Large \bfseries Supplementary Material}

\vspace{1em}
{\large A Realistic Protocol for Evaluation of Weakly Supervised Object Localization}
\end{center}

\vspace{2em}
]

\maketitle

\input{sections/2_supp_content}

\end{document}

%% file: aux_files/macros.tex
\usepackage{xcolor}

\newcommand\cub{\texttt{CUB-200-2011}\xspace}
\newcommand\cubs{\texttt{CUB}\xspace}

\newcommand\ilsvrc{\texttt{ILSVRC}\xspace}
\newcommand\ilsvrcvtwo{\texttt{ILSVRC-V2}\xspace}

\newcommand\glas{\texttt{GLaS}\xspace}
\newcommand\camelyon{\texttt{CAMELYON}\xspace}

\newcommand\maxboxacc{\texttt{MaxBoxAcc}\xspace}
\newcommand\newmaxboxacc{\texttt{MaxBoxAccV2}\xspace}
\newcommand\newmaxboxaccs{\texttt{MaxBoxV2}\xspace}

\newcommand\pxap{\texttt{PxAP}\xspace}

\newcommand\cls{\texttt{CLS}\xspace}
\newcommand\classification{\texttt{CL}\xspace}
\newcommand\localization{\texttt{LOC}\xspace}

\newcommand\iou{\texttt{IoU}\xspace}
\newcommand\iouthirty{\texttt{IoU-30}\xspace}
\newcommand\ioufifty{\texttt{IoU-50}\xspace}
\newcommand\iouseventy{\texttt{IoU-70}\xspace}
\newcommand\red[1]{{\color{red}#1}}

\definecolor{darkergreen}{RGB}{21, 152, 56}
\definecolor{red2}{RGB}{252, 54, 65}
\definecolor{green2}{RGB}{1, 129, 1}
\definecolor{Gray}{gray}{0.85}
\newcommand\green[1]{{\color{green2}#1}}

\newcommand\btst{\texttt{BT-TT}\xspace}
\newcommand\bval{\texttt{BV-TT}\xspace}
\newcommand\btestvalth{\texttt{BT-VT}\xspace}
\newcommand\bvalvalth{\texttt{BV-VT}\xspace}
\definecolor{btst}{rgb}{0.9, 1.0, 0.9} 
\definecolor{bval}{rgb}{0.9, 0.9, 1.0} 
\definecolor{btestvalth}{rgb}{1.0, 0.9, 0.9} 
\definecolor{bvalvalth}{rgb}{1.0, 1.0, 0.9} 

\newcommand{\fakeslant}[2][0.25em]{\makebox[2.5pt][l]{#2}\kern-#1#2}
\newcommand{\boldcourier}[1]{\textbf{\texttt{\fakeslant{#1}}}}

\newcommand\backbonebold{\boldcourier{\fontfamily{lmtt}\fontseries{b}\selectfont Backbone}\xspace}

\newcommand\gt{\texttt{GT}\xspace}
\newcommand\cliplabels{\texttt{CLIP}\xspace}
\newcommand\rpnlabels{\texttt{RPN}\xspace}
\newcommand\sslabels{\texttt{SS}\xspace}

%% file: sections/0_abstract_content.tex
\vspace{-0.2cm}
Weakly Supervised Object Localization (WSOL) allows training deep learning models for classification and localization (\localization) using only global class-level labels. The absence of bounding box (bbox) supervision during training raises challenges in the literature for hyper-parameter tuning, model selection, and evaluation. WSOL methods rely on a validation set with bbox annotations for model selection, and a test set with bbox annotations for threshold estimation for producing bboxes from localization maps. This approach, however, is not aligned with the WSOL setting as these annotations are typically unavailable in real-world scenarios. Our initial empirical analysis shows a significant decline in \localization performance when model selection and threshold estimation rely solely on class labels and the image itself, respectively, compared to using manual bbox annotations. This highlights the importance of incorporating bbox labels for optimal model performance. In this paper\footnote{\noindent\textbf{Our scope.} This work focuses on WSOL~\cite{belharbi2022fcam,murtaza2023dips, Murtaza_2023_WACV, murtaza2022constrained}, as opposed to other related tasks such as weakly supervised detection, segmentation, or instance segmentation, which are often mixed in earlier works \cite{WSOLasMIL2,WSOLasMIL3,OldWSOL1}.}, \textbf{a new WSOL evaluation protocol} is proposed that provides \localization information without the need for manual bbox annotations. In particular, we generated noisy pseudo-boxes from a pretrained off-the-shelf region proposal method such as Selective Search, CLIP, and RPN for model selection. These bboxes are also employed to estimate the threshold from \localization maps, circumventing the need for test-set bbox annotations. Our experiments\footnote{Our code and generated pseudo-bounding boxes can be accessed at \href{https://github.com/shakeebmurtaza/wsol_model_selection}{github.com/shakeebmurtaza/wsol\_model\_selection}. \vspace{-0.17cm}} with several WSOL methods on challenging natural and medical image datasets show that using the proposed pseudo-bboxes for validation facilitates the model selection and threshold estimation, with \localization performance comparable to models selected using GT bboxes on the validation set and threshold estimation on the test set. It also outperforms models selected using class-level labels, and then dynamically thresholded with only \localization maps. 


%% file: sections/1_main_content.tex
\vspace{-0.6cm}
\section{Introduction} \label{wacveval:sec:intro}
\vspace{-0.15cm}

\begin{figure}[!b]
     \centering
    \vspace{-0.2cm}
    \begin{tabular}{@{} c c @{}}
        \begin{minipage}[l]{0.015\textwidth}
            \vspace{-0.9cm}\subcaption{}\label{wacveval:fig:oneconvergcurvesing}
        \end{minipage} &
        \begin{minipage}[c]{0.44\textwidth}
            \centering
            \includegraphics[width=\textwidth]{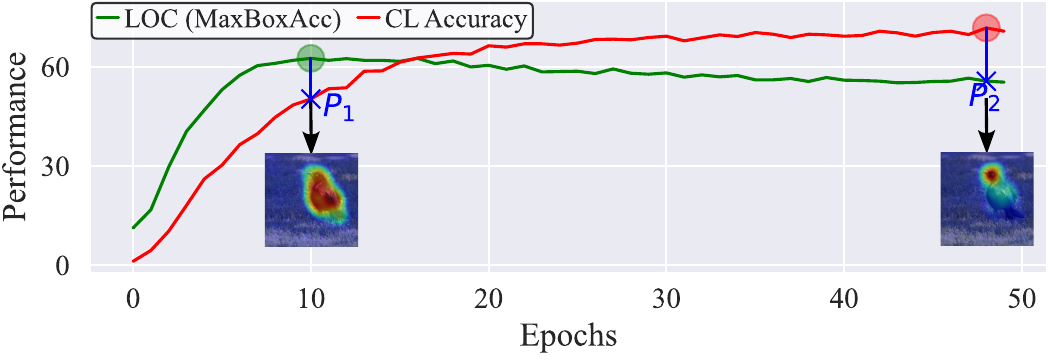}
        \end{minipage} \\
        \begin{minipage}[c]{0.015\textwidth}
            \vspace{-0.9cm}\subcaption{}\label{wacveval:fig:modeSelecclloc}
        \end{minipage} &
        \begin{minipage}[c]{0.44\textwidth}
            \centering
            \includegraphics[width=\textwidth]{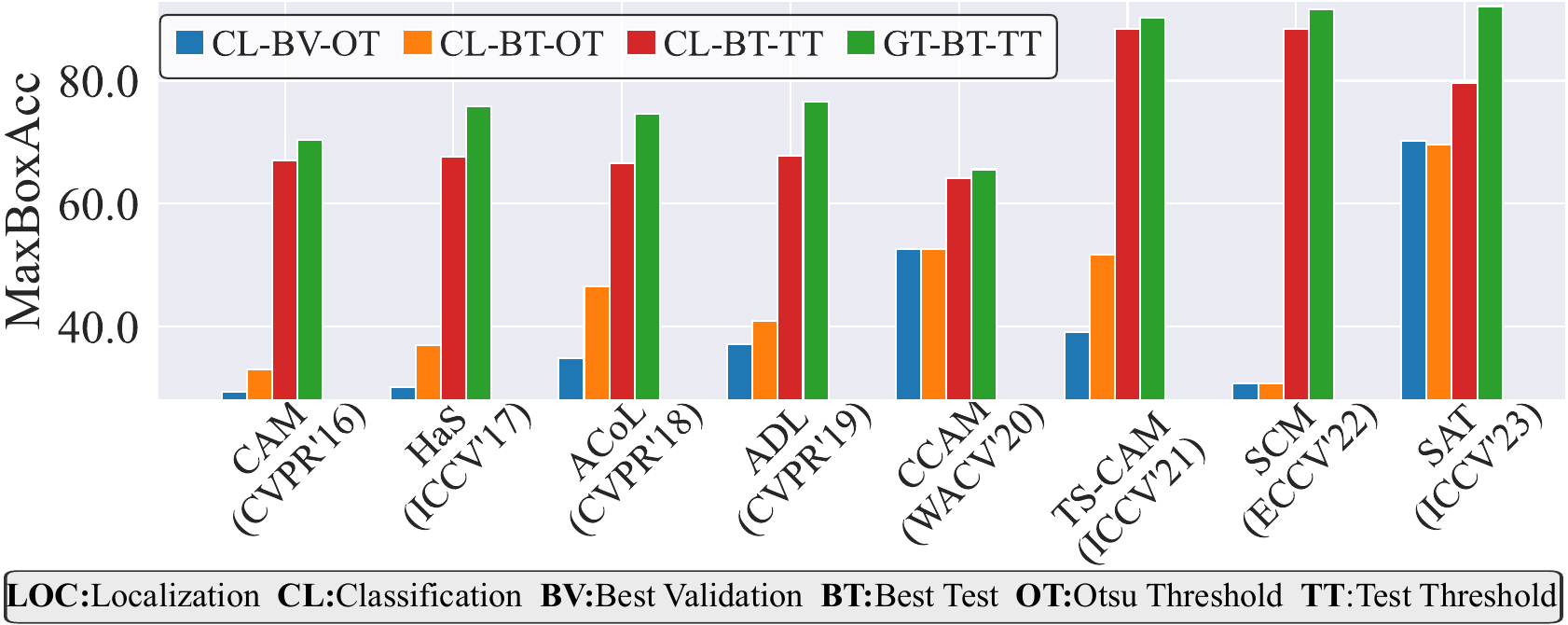}
        \end{minipage}
    \end{tabular}
        \caption{
        \textbf{(A)} Example of \localization (\green{\maxboxacc}) and \red{\classification} accuracy w.r.t the number of epochs on the \cubs validation set in a WSOL setting, where only class-labels are used to train the model.
        These curves show that \localization and \classification tasks are loosely correlated where convergence is reached at different training epochs. Typically, high \localization performance is achieved early in the training, followed by degradation. However, the classifier takes a longer time to converge. Selecting the best model for \localization may need a \localization annotation in the validation set.  
        \textbf{(B)} Significant bias arises when using GT bounding boxes and test set thresholds, leading to overestimated \localization accuracy, as bbox annotations are typically absent in real-world scenarios. \textit{In a realistic WSOL scenario (CL-BV-OT), where model selection is based on \classification accuracy and OT over the validation set, performance declines considerably.}
        }
        \label{wacveval:fig:issue-model-selection}
\end{figure}

Methods for WSOL provide efficient learning strategies~\cite{bai2022weakly,gao2021ts,wuspatial23} to mitigate the cost of annotating datasets with bboxes. Given only coarse class-labels, a deep learning model can be trained to perform image classification (\classification) tasks while yielding the spatial image region of the object (localization). Class-Activation Mapping (CAM)~\cite{choe2020evaluating,rony2023deep} is the dominant WSOL approach that builds a spatial heatmap to localize an object.

Despite its popularity, WSOL still faces several challenges, in particular with its evaluation protocol~\cite{choe2020evaluating}. We argue that the common protocol presently used in the literature may not be reliable for real-world applications limiting WSOL benefits. WSOL methods should only require class-labels for supervision. During training (with held-out validation sets), the model should not have access to data with manual localization (\localization) supervision, i.e., ground truth (GT) bounding box (bbox) annotations. Model selection and hyper-parameter search are therefore very challenging (Fig.\ref{wacveval:fig:oneconvergcurvesing}). Early works observed the test performance for hyper-parameter selection leading to an overestimation of model performance~\cite{choe2019attention,singh2017hide,zhang2018adversarial}\footnote{Refer to the official implementation~\cite{choe2019attention,singh2017hide,zhang2018adversarial} 
that is publicly available. Unfortunately, even recent works after~\cite{choe2020evaluating} still use the test set as a held-out validation set~\cite{wuspatial23} as shown in their public code. This practice potentially biases the selection process in favor of \localization performance.\vspace{-0.1cm}}. As illustrated in Fig.\ref{wacveval:fig:modeSelecclloc}, the test-set \localization performance of these methods may decline in real-world scenarios, where only class-labels are available.

Choe et al.\cite{choe2020evaluating} proposed an improved evaluation protocol to address some of the above limitations. To perform model selection, they propose using a held-out validation set with full manual annotations (class labels and bboxes). Although this protocol allows us to avoid using the test set and to perform unbiased modal evaluations, it is somewhat unrealistic since annotated bboxes are not available in practice. Having access to such manual \localization supervision during training goes against the WSOL setting, and can overestimate model performance (Fig.\ref{wacveval:fig:modeSelecclloc}). Moreover, the same authors~\cite{choe2020evaluating} have shown that a model directly trained on those annotated images would provide higher \localization accuracy than any weakly supervised approach. Although authors annotated a limited number of images per class to create a held-out validation set, the total cost of annotation adds up with the number of classes. For instance, they annotated 10 images/class for the \ilsvrc dataset, which amounts to a total of 10,000 images. Given the cost, these fully annotated images can, for example, be directly used to fit the model weights in a semi-supervised learning (SSL) setting rather than being used for model selection in a WSOL setting.

Another unrealistic practice prevalent in the literature is the use of GT bboxes from the test set to estimate the threshold for binarizing the \localization map to generate bboxes~\cite{choe2020evaluating}. However, this practice leads to an overestimation of model performance (\localization accuracy) because bbox annotations are typically absent in real-world scenarios, making it impossible to estimate the threshold during inference. Models selected using GT bboxes and test set thresholds may exhibit significantly higher \localization accuracy than models selected based on \classification accuracy and thresholds automatically estimated using the Otsu thresholding (OT) algorithm, for example, as one would do when bbox annotations are unavailable on the validation. 
An example of the inflated performance resulting from using the test set for model selection and threshold estimation is illustrated in Fig.\ref{wacveval:fig:modeSelecclloc}, where the \localization accuracy is reported for various WSOL methods on the CUB test set using two model selection criteria: \classification and \localization accuracy. Additionally, the results of different thresholding methods are reported for \localization maps used to generate bboxes. Additionally, we presented the test \localization accuracy when the model was selected with the best \classification accuracy on the validation or test set, and when automatic thresholding algorithm (OT) was used for producing bboxes.
In a more realistic evaluation protocol, where model selection is based on \classification accuracy and the OT on the validation set (blue bar Fig.\ref{wacveval:fig:modeSelecclloc}), a noticeable decrease in performance is observed. This assessment highlights the necessity for unbiased thresholding and model selection to evaluate the real-world capabilities WSOL of methods.

This paper provides a realistic evaluation protocol for WSOL that can be applied in real-world scenarios. Instead of assuming that manual GT bbox annotations are available for validation, we estimate pseudo-bboxes by leveraging an off-the-shelf pretrained model that can generate region-of-interest (ROI) proposals~\cite{Hosang14} such as Contrastive Language–Image Pre-training (CLIP)~\cite{lin23} (Fig.\ref{wacveval:fig:proposedportocol}). This model is pretrained on generic datasets requiring less supervision, such as class-level labels. We also consider Selective Search (SS)~\cite{uijlings13}, an unsupervised method that does not require training, and Region Proposal Networks (RPN)~\cite{ren15}, trained on large datasets with class-agnostic bboxes. They can generate multiple bboxes with objectness scores for ranking. The impact of using these different levels of supervision is explored for pretrained proposal generators.
Given a set of proposed bboxes for an image, and using pointing game analysis~\cite{zhang2018top} and class-level labels, the most discriminative bboxes are selected and ranked using classifier response to produce the final bbox. This approach allows to create pseudo-annotated bboxes for the held-out validation set, and therefore, avoids manual annotation, leading to more realistic model selection. Despite their lower accuracy, these pseudo-bboxes can effectively select models and enable the estimation of thresholds using the validation set for generating bboxes from \localization maps in the test set.

To address the biased estimation obtained with current approaches, a threshold is estimated using the validation-set pseudo-boxes and subsequently applied to the test set to convert \localization maps into bboxes. Estimating the threshold from the validation set enables WSOL methods to achieve realistic and competitive performance in real-world scenarios. This avoids estimating the threshold from the test set, which leads to an overestimation of performance, and using OT which leads to low performance (see Fig.\ref{wacveval:fig:modeSelecclloc}).

\noindent\textbf{Our main contributions are summarized as follows.} \\
\noindent\textbf{(1)} An initial experimental analysis of state-of-the-art WSOL methods on the \cubs and \ilsvrc datasets showing the impact on the test-set performance of bbox supervision in the held-out validation set. Results indicate that utilizing only class-labels for validation leads to poor test-set \localization accuracy, highlighting the importance of using \localization cues for model selection. Conversely, using manually annotated bboxes for validation improves \localization accuracy, leading to an overestimated and unrealistic performance evaluation. Further experiments evaluating the use of less precise pseudo-bbox supervision in the validation set indicate that noisy bboxes can still effectively guide model selection. \\
\noindent\textbf{(2)} A realistic evaluation protocol is proposed for WSOL. First, we propose to estimate pseudo-bboxes without manual intervention and use only class-labels to generate annotations for a held-out validation set. Given different application scenarios, we explored three different bbox generation strategies based on off-the-shelf models: SS~\cite{uijlings13}, RPN~\cite{ren15}, and CLIP~\cite{lin23}. Secondly, these pseudo-bboxes generated for the validation set allow for estimation of the threshold, which can be subsequently applied to test images for producing bboxes on the \localization maps. \\
\noindent\textbf{(3)} An extensive set of experiments was conducted with prominent WSOL methods using our new protocol on the \cubs and \ilsvrc datasets for model selection and threshold estimation to convert localization maps into bboxes. To assess the generalizability of our model, our analysis was extended to include \glas and \camelyon medical image data (see the supplementary material). Results show that models selected and thresholds estimated using pseudo-bboxes achieve a level of accuracy consistent with models selected using GT bboxes (currently-used but unrealistic selection protocol). Furthermore, using a single threshold estimated from pseudo-boxes reduced the threshold search space from 1,000 values to just one.

\section{Related Work} \label{wacveval:sec:related-w}
In supervised learning, hyper-parameter and model selection typically rely on a fully annotated validation set. Methods like early stopping leverage this set for realistic estimation of model performance on unseen test data~\cite{Finnoff93,Lodwich09,morgan89} since the three sets (train, val and test) employ the same supervisory signal and perform the same task. This section reviews model selection protocols in different settings, highlighting issues in model selection and hyperparameter search. It reviews early stopping methods without a validation set and concludes with model selection in WSOL.

\noindent\textbf{(a) Protocols with limited supervision.} Beyond the fully supervised setting, other learning settings have been derived with their protocol. This typically concerns learning with less annotated data~\cite{cheplygina19,zhou18}. This may affect the availability of annotations over the held-out set used for model selection and hyper-parameter search. Despite the efforts of the community to design standardized protocols, unrealistic practices may emerge. Different attempts have been made to ensure that such protocols are realistic to benefit from the unbiased application in real-world scenarios.

In the SSL setting, researchers have investigated whether exiting works can perform well on realistic benchmarks compared to standard benchmarks used by the community~\cite{Oliver18,Su21}. The authors argue that standard SSL benchmarks fail to reflect real-world challenges such as sensitivity to unlabeled data size, distribution shifts, class imbalance, and novel classes, and reserving some labeled samples as a held-out validation set reduces the training set size.

Unsupervised domain adaptation (UDA) is another learning setting with less annotation that raises several questions about how realistic is the protocol. UDA involves adapting a source model on labeled source data and unlabeled target data. Model selection and hyper-parameter search are challenging, with recent works improving evaluation protocols~\cite{Dinu23,Ericsson23,Saito21,salvador22,Yang23,You19}. For instance, Ericsson et al.~\cite{Ericsson23} show that using an annotated target set for hyper-parameter search leads to largely overestimated and unrealistic performance. They proposed using source data, evaluating another UDA model such as InfoMax~\cite{Shi12}, or a labeled held-out target set. Yang et al.\cite{Yang23} introduced the Transfer Score for UDA model evaluation for hyper-parameter search and early stopping. The protocol issue is much more severe in the case of source-free UDA~\cite{fang23} where the user cannot access the source data.

Self-supervised~\cite{caron21}, unsupervised learning~\cite{chen23,vaithyanathan99}, and data generation~\cite{Borji19} also face model selection and hyper-parameter search difficulties due to limited supervision. For instance, in image generation, quantitative measures~\cite{Heusel17,Salimans16,Zhang18} may not fully capture how realistic a generated image is, necessitating visual inspections that affect reproducibility and real-world applicability.

The aforementioned works show that the evaluation protocol is an issue for many settings. This work focuses on realistic WSOL evaluation protocols, in particular model selection for hyperparameter optimization and threshold estimation using test set for generating bboxes. 

\noindent\textbf{(b) Early stopping without validation set.}
Some works have explored early stopping without a validation set, crucial when data is limited. Large held-out validation sets improve generalization estimates but reduce training data size at the cost of significantly reducing the data available for the training set. Mahsereci et al.\cite{Mahsereci17} proposed a criterion using gradient statistics over the training set, effective in classification and regression tasks. Duvenaud et al.\cite{duvenaud16} used weight statistics, monitoring the entropy change in network parameters as an indicator of generalization. Yuan et al.\cite{yuan24} introduced \emph{Label Wave}, tracking model prediction changes on noisy training labels to prevent overfitting. Li et al.\cite{li20} suggested using augmented training data instead of a real validation set to reduce computational costs. While these works aim to halt the training of the entire model, Bonet el al.\cite{Bonet21} per-class channel early stopping for CNNs without validation sets, leveraging DeepNNK\cite{Shekkizhar20}, a non-negative kernel regression.
Note that in these works, the training and test task is the same. In WSOL, however, the model is trained for the \classification task but it is evaluated on the \localization task.

\noindent\textbf{(c) Protocols for WSOL.} Before the work of Choe et al.~\cite{choe2020evaluating}, bbox annotations were not considered to create the held-out validation set. Therefore, practitioners may unintentionally inspect the test \localization performance, leading to unrealistic WSOL~\cite{choe2019attention,singh2017hide,zhang2018adversarial}. To alleviate this, Choe et al.~\cite{choe2020evaluating} investigated several aspects of WSOL protocol. They mainly showed that state-of-art methods (at that time) do not improve accuracy beyond that of the baseline CAM method~\cite{zhou2016learning} after searching for the optimal decision threshold. They proposed a concise WSOL evaluation protocol, including new metrics and a held-out validation with bboxes. Although their work largely improved the WSOL protocol, it is somewhat unrealistic since manually-labelled bboxes are not available in real-world applications of the WSOL setting. Strong supervision is used to select a model, which implicitly helps to select a model with the best \localization performance which can contribute to overestimating \localization accuracy on the test set.
In addition, it has been reported~\cite{choe2020evaluating} that using manually annotated bboxes for model training on the held-out validation set, rather than class-labels on the training set, can outperform baseline WSOL methods. Furthermore, Choe et al.~\cite{choe2020evaluating} also employ the test set for estimating the threshold needed to convert the localization map to bboxes, rendering WSOL setting unrealistic. 

Although more realistic protocols have been developed for other learning settings like SSL and UDA, WSOL represents a challenging setting because of the different tasks (\classification and \localization) performed with the same model during training vs testing.  
The supervision available during training allows us to select models for the \classification task. At test time, the model is evaluated additionally over \localization. The disconnection between available annotation at training time, and the required task at test time creates a challenging situation for model selection. 
Furthermore, $\S$\ref{wacveval:sec:realisticprtocol} provides a more realistic evaluation protocol within the WSOL framework.

\section{A Realistic Protocol for Model Evaluation}
\label{wacveval:sec:realisticprtocol}

In this section, we describe a new protocol for evaluating WSOL methods. First \localization accuracy is shown to be overestimated when using manual bbox supervision for the held-out validation set. To address this, automatically generating pseudo-bboxes is proposed for the validation set using only class-label supervision, eliminating the need for manual annotation. It involves generating pseudo-bboxes for the validation set using off-the-shelf ROI proposal generators and utilizing these pseudo-bboxes to estimate a threshold. Then this can be applied to localization maps in the test set for producing bboxes, providing a more systematic and realistic evaluation protocol.

\subsection{Model Selection}
Consider a WSOL training set ${\mathcal{D}_{train}=\{(x_i, y_i, \cdot)\}}$ composed of samples with ${x_i}$ the input image, ${y_i}$ its class-label, and ${\cdot}$ is its missing bbox supervision. Let us define ${\mathcal{D}_{val}=\{(x_i, y_i, \cdot)\}_{i=1}^N}$\footnote{In Choe et al.~\cite{choe2020evaluating} the validation set contain ${o(x_i)}$.}, and ${\mathcal{D}_{test}=\{(x_i, y_i, o(x_i))\}}$ as the held-out validation and test sets, respectively, where ${o(x_i)}$ is the oracle manual bbox annotation for the sample ${x_i}$. In the WSOL setting, bbox supervision is unavailable for any dataset. However, in a laboratory, the test set contains them to evaluate \localization performance of a model. 

The WSOL setting involves training a model ${f(\cdot; \theta)}$ with parameters ${\theta}$ to correctly classify an input image while yielding \localization boxes. Training is performed on ${\mathcal{D}_{train}}$ that lacks \localization supervision. In this paper, we introduce a pseudo-annotator ${\hat{o}}$ that automatically annotates an image with pseudo-bboxes (without human intervention) and only requires the class-labels. ${\hat{o}}$ yields noisy bboxes, providing \localization pseudo-bbox annotations ${\hat{o}(x_i)}$ that are less accurate than oracle bbox annotations ${o(x_i)}$. Moreover, we redefine the validation set $\mathcal{D}_{val}$ to incorporate pseudo-bboxes ${\hat{o}(x_i)}$, yielding ${\mathcal{D}_{val}=\{(x_i, y_i, {\hat{o}(x_i)})\}_{i=1}^N}$.

Let ${f(x;\theta)}$ be the model's \classification prediction, while  ${f(x;\theta)^b}$ is the predicted bbox for image $x$. We denote by ${\ell(\cdot, \cdot)}$ a \localization similarity measure between two bboxes. Following Choe et al.~\cite{choe2020evaluating}, ${\ell(\cdot, \cdot)}$ is commonly \maxboxacc or simply the Intersection-Over-Union (\iou) measure. Let us define a hypothetical ideal scenario where one can have access to the oracle for fully annotated bboxes ${o(x_i)}$ in the validation set ${\mathcal{D}_{val}}$ allowing, for instance, to perform early stopping and threshold selection. Therefore, we can define the average \localization performance of the model $f$ with a specific parameter value of $\theta$ over this validation set as,
\begin{equation}
    \label{wacveval:eq:vl_exact_sup}
    \mathcal{M}(\theta, \mathcal{D}_{val})_o = \frac{1}{N} \sum_i^N \ell(f(x_i;\theta)^b, o(x_i)).
\end{equation}
Similarly, the same measure over ${\mathcal{D}_{val}}$ can be defined using our pseudo-annotator ${\hat{o}}$ as,
\begin{equation}
    \label{wacveval:eq:vl_noisy_sup}
    \mathcal{M}(\theta, \mathcal{D}_{val})_{\hat{o}} = \frac{1}{N} \sum_i^N \ell(f(x_i;\theta)^b, \hat{o}(x_i)).
\end{equation}

\begin{figure}[!b]
    \centering
    \vspace{-0.4cm}
    \includegraphics[width=1\linewidth]{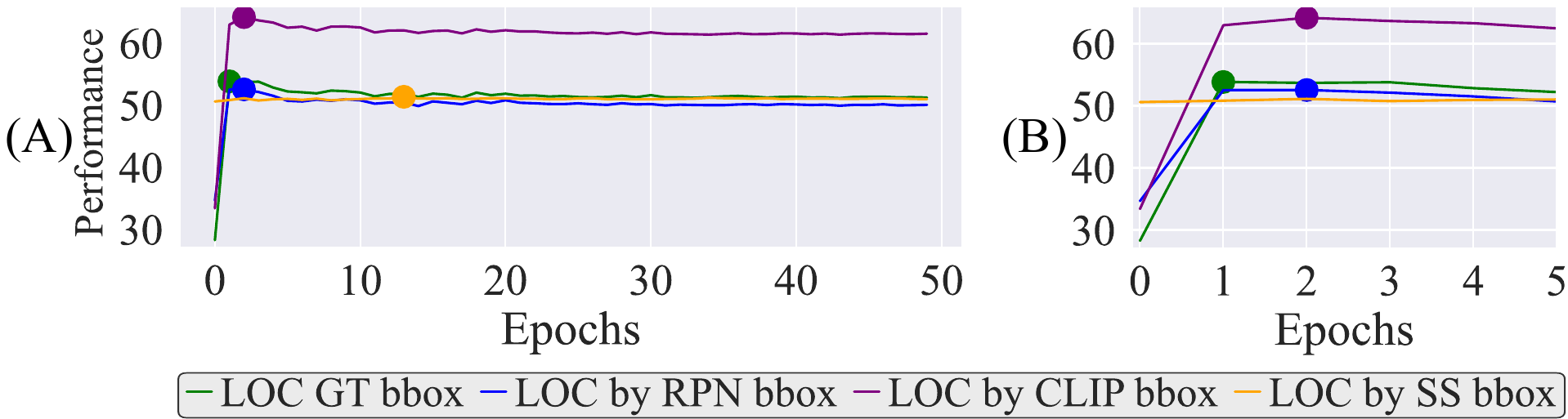}
    \caption{Model selection with early stopping at the epoch indicated with a dot using \localization accuracy. Different approaches are compared for pseudo-bbox annotation versus GT annotations (oracle). Fig(B) is a zoom of Fig.(A) between epochs 0 and 5.
    Results are reported over \cubs validation set using the CAM method~\cite{zhou2016learning} with \iou as a \localization measure. \localization curves with pseudo-bbox annotations typically have similar behaviour to the Oracle GT bbox annotations, making them suitable for WSOL model selection. They increase and reach their peak for a similar number of epochs, followed by a decline and stagnation in performance. In contrast, the \classification curve reaches its peak toward the end of training when \localization performance has already degraded. \classification accuracy is therefore inadequate as a WSOL selection criterion to achieve high \localization accuracy performance. Misalignment between \localization and \classification behaviour has been studied further in~\cite{choe2020evaluating}. 
    \vspace{-0.19cm}
    }
    \label{wacveval:fig:behviour_curves}
\end{figure}

Eqs. \ref{wacveval:eq:vl_exact_sup} and \ref{wacveval:eq:vl_noisy_sup} provide an assessment of the model \localization accuracy over a held-out ${\mathcal{D}_{val}}$. Eq.\ref{wacveval:eq:vl_exact_sup} is accurate, while Eq.\ref{wacveval:eq:vl_noisy_sup} accounts for errors due to inaccurate \localization annotation. Since practitioners do not have access to the oracle bboxes ${o(x_i)}$, the measure ${\mathcal{M}(\theta, \mathcal{D}_{val})_o}$ is not realistic. When one has access to oracle ${o(x_i)}$ bboxes for training, it is better to perform SSL to directly fit the model weights instead of WSOL~\cite{choe2020evaluating}. However, one can still assess ${\mathcal{M}(\theta, \mathcal{D}_{val})_{\hat{o}}}$ for an approximate but realistic estimation of the model's \localization accuracy. The latter measure provides a realistic assessment of \localization accuracy for WSOL. A direct application of our proposed noisy measure ${\mathcal{M}(\theta, \mathcal{D}_{val})_{\hat{o}}}$ in the WSOL setting is early stopping for model selection, and hyper-parameter search. In these applications, the measure must follow a similar behaviour to ${\mathcal{M}(\theta, \mathcal{D}_{val})_o}$, as shown in Fig.\ref{wacveval:fig:behviour_curves}. 

\noindent\textbf{Pseudo-bbox annotation}. For generation of pseudo-bboxes (using ${\hat{o}}$), we leverage pretrained of-the-shelf region proposal models to generate \localization annotations for a held-out validation set ${\mathcal{D}_{val}=\{(x_i, y_i, {\hat{o}(x_i)})\}_{i=1}^N}$. This set is originally labeled with class-labels only. These models can be unsupervised without the need for training or require weak supervision for training. We consider the three following scenarios of the pretrained model depending on its level of supervision. 

\noindent\emph{(a) Unsupervised}: In this case, the model does not have access to any supervision. Typical examples are conventional region proposal methods~\cite{Hosang14} that are generally used to build pseudo-supervision to train object detector models~\cite{meethal22}. SS~\cite{uijlings13} is a commonly used method where no parameters are learned. Carefully engineered features and score functions are used to greedily merge low-level superpixels that produce the final region proposals. The per-proposal score can be used to rank the likelihood that the proposal contains an object.

\noindent\emph{(b) Supervised with class-labels}: A middle-ground approach is to use class-labels which are available in the WSOL setting. The CLIP model~\cite{lin23} is a model trained on millions of generic text-image pairs from the Internet. Pretrained CLIP is used to pseudo-annotate our held-out $\mathcal{D}_{val}$. Class labels are used as a prompt to generate a CAM for the labeled object in the image. The CAM is automatically thresholded using the Otsu method~\cite{otsuthresh79}. Then, a bbox is constructed around each connected object. Since CLIP points to the discriminative region associated with the input class-label, we select the largest bbox as a pseudo-bbox.

\noindent\emph{(c) Supervised with class agnostic bboxes}: An alternative approach involves using a region proposal model that is trained using a general dataset like Pascal VOC and MS-COCO, with \localization bboxes. This methodology differs in that it does not employ class-labels, but rather \localization bboxes that are class-agnostic. A typical example is the RPN~\cite{ren15}, a fully convolutional network that simultaneously predicts object bounds at each position, and provides objectness scores that allow to rank proposals.

\begin{figure}[!b]
\vspace{-0.35cm}
\begin{center}
\includegraphics[width=1\linewidth,trim=0 0 0 0, clip]{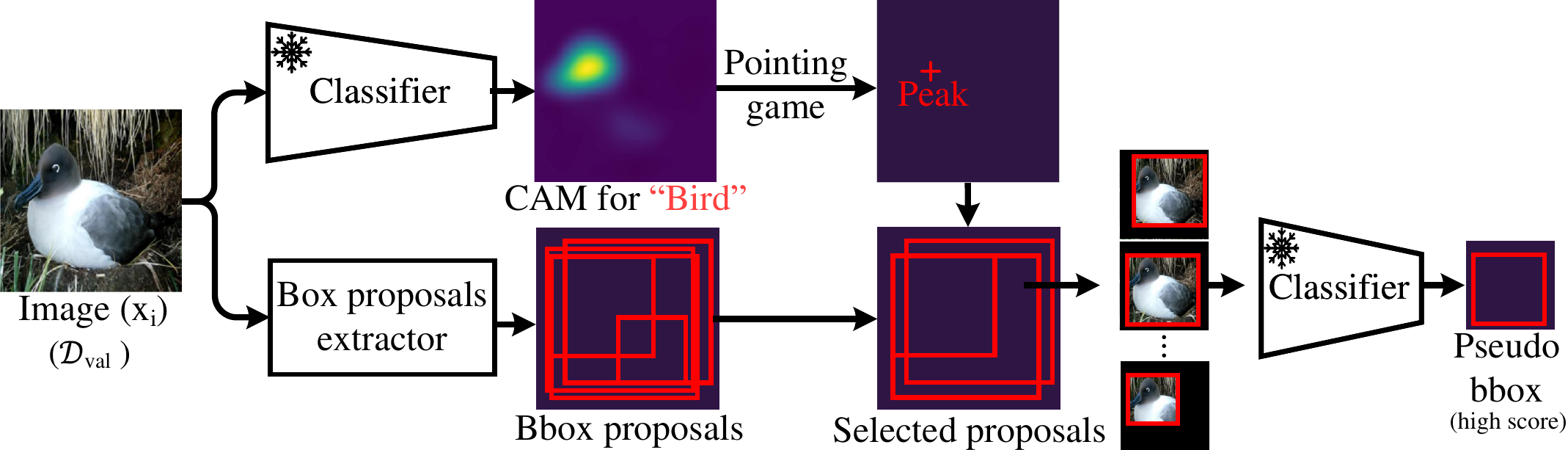}
\end{center}
\vspace{-0.47cm}
\caption{Our proposed \localization pseudo-bbox annotator ${\hat{o}}$. A set of bbox proposals is initially extracted using a region proposal model, and from them, discriminative ones are selected using pointing game analysis~\cite{zhang2018top}. In the case where multiple bboxes are selected, the classifier confidence over the foreground region is used to select the most discriminative bbox.
\vspace{-0.25cm}
}\label{wacveval:fig:proposedportocol}
\end{figure}
 
For SS and RPN several bboxes are generated per image. To discard irrelevant bboxes and select the most discriminative one, we leverage the pointing game analysis~\cite{zhang2018top} (see Fig.\ref{wacveval:fig:proposedportocol}). To this end, a classifier pretrained over ${\mathcal{D}_{train}}$ is considered. It is trained until convergence using only class-labels. The pointing game uses the maximum CAM response to select the most discriminative bboxes. In the case where multiple bboxes are ultimately selected, they are scored by the classifier response, and the bbox with the highest score is ultimately selected. Pseudo-bbox annotations are constructed for ${\mathcal{D}_{val}}$, where each box contains the most discriminative object labeled in the image. This process is only executed once before performing any WSOL training. The generated bboxes are stored on disk, and used for future WSOL training. Therefore, our approach does not add any computation time to the training itself.

Fig.\ref{wacveval:fig:behviour_curves} shows the behaviour of \localization performance when using pseudo-bboxes vs oracle bboxes over ${\mathcal{D}_{val}}$. The overall trend is similar as they both increase during the first epochs and reach their peak, decline and stagnate over the following epochs. This is extremely helpful for early stopping and hyper-parameter search. We also observe that \localization using CLIP pseudo-annotation yields higher \localization than the with the oracle. This is plausible since a model prediction can be further aligned with some annotations than others.

To provide a realistic evaluation protocol for WSOL, we proposed a strategy to generate pseudo-bboxes for ${\mathcal{D}_{val}}$. It bypasses the need for manually annotated (oracle) bboxes by leveraging pretrained region proposal models. Then, the pointing game analysis is used to select the most discriminative proposals. In $\S$\ref{wacveval:sec:res-discussion}, we show that using manually annotated bboxes for ${\mathcal{D}_{val}}$ set can increase \localization accuracy on ${\mathcal{D}_{test}}$, leading to performance overestimation. We also show that noisy pseudo-bboxes can effectively be used for model selection in the WSOL. This is initially shown using synthetic boxes, then using our generated pseudo-bboxes. These pseudo-bboxes are made public to the researchers for the design of more realistic WSOL methods.

\subsection{Threshold Estimation}
WSOL methods seek to produce bboxes corresponding to particular objects. They generate a \localization map $\mathcal{M}=\mathcal{M}^c\in\mathbb{R}^{H,W}$ highlighting the object of a class $c$, where $\mathcal{M}_{ij}$ denotes the pixel value at location $(i,j)$ for $i\in\{1,2,\ldots,H\}$ and $j \in \{1, 2, \ldots, W\}$. Higher values of $\mathcal{M}_{ij}$ indicate the foreground and lower values indicate the background. The \localization map $\mathcal{M}$ is first normalized due to varying score statistics across images. After normalization, WSOL methods apply a threshold $\tau\in[0,1]$ to $\mathcal{M}$ to create a binary mask $\{\mathcal{M}_{ij} \geq \tau\}$ for producing bbox. Conventionally, researchers either fix $\tau$ or tune it as a hyperparameter. However, the optimal $\tau$ varies based on the architecture and dataset. Choe et al. \cite{choe2020evaluating} proposed using multiple thresholds, $\tau \in [0,1]$, to compute bboxes for different masks, and selecting the optimal threshold $\tau^*$ based on performance statistics. While effective, this method tends to overestimate performance by using ground-truth bboxes for threshold estimation, which is unrealistic in the WSOL setting. Moreover, for $\tau$ estimation, different automatic thresholding methods like OT
\footnote{OT evaluates all possible threshold values, calculates the variance within each of the two clusters (background and foreground), and selects the threshold that minimizes the weighted sum of these variances.} 
lead to an underestimation of performance, as they fail to consider the contextual relationships within the image. To address these issues, we propose using pseudo-bbox labels ${\hat{o}(x_i)}$ produced by our method for $\tau$ estimation. This approach is more realistic and achieves performance that is competitive to the case where $\tau$ is estimated with the test set.
\vspace{-0.1cm}
 
\section{Experimental Validation}\label{wacveval:sec:res-discussion}
\subsection{Methodology}\label{wacveval:sec:exp-meth}
\vspace{-0.1cm}
\noindent\textbf{Datasets.}
We employed well-established and challenging datasets that are commonly used for evaluating WSOL models. They are, \textbf{(i) \cubs} \cite{wah2011caltech}, encompasses 200 classes with a total of 11,788 images. These are divided into 5,994 images for training and 5,794 for testing. Additionally, for validation, we employed an independent validation set comprising 1,000 images representing five per class, as provided by \cite{choe2020evaluating}. \textbf{(ii) \ilsvrc}
~\cite{imagenet_cvpr09} consists of around 1.2 million images for training and 10,000 images for the validation set, distributed across 1,000 classes. We utilized the validation split included with the original dataset as our test set, given its ample sample size suitable for testing. However, for validation, we utilized \textbf{\ilsvrcvtwo}, collected by \cite{recht2019imagenet} and annotated by \cite{choe2020evaluating}, which minimizes biases toward the test set. Moreover, to ensure a consistent and fair comparison, we closely adhered to the dataset-splitting strategy proposed by \cite{choe2020evaluating} for both datasets. \textbf{(iii)} Extended results on two medical datasets (\glas and \camelyon) are presented in supplementary material.

\noindent\textbf{WSOL methods.}
To compare the performance of the proposed protocol, we employed eight state-of-the-art WSOL methods, which include CAM~\cite{zhou2016learning}, HaS~\cite{singh2017hide}, ACoL~\cite{zhang2018adversarial}, ADL~\cite{choe2019attention}, NL-CCAM~\cite{yang2020combinational}, TS-CAM~\cite{gao2021ts} SCM~\cite{bai2022weakly}, and SAT~\cite{wuspatial23}. Further details regarding these methods can be found in the supplementary material.

\renewcommand{\tabcolsep}{1pt}
\begin{figure*}[!t]
\vspace{-0.2cm}
    \centering
    \begin{tabular}{@{} c c @{} @{} c c @{} @{} c c @{}}
        \multirow{2}{*}{
        \begin{minipage}[l]{0.015\textwidth}
            \vspace{.9cm}\subcaption{}\label{wacveval:fig:noislevel}
        \end{minipage}} &
        \multirow{2}{*}{
        \begin{minipage}[c]{0.3\textwidth}
            \vspace{-0.6cm}
            \centering
            \includegraphics[width=1\textwidth,trim=0 0 0 0cm,clip]{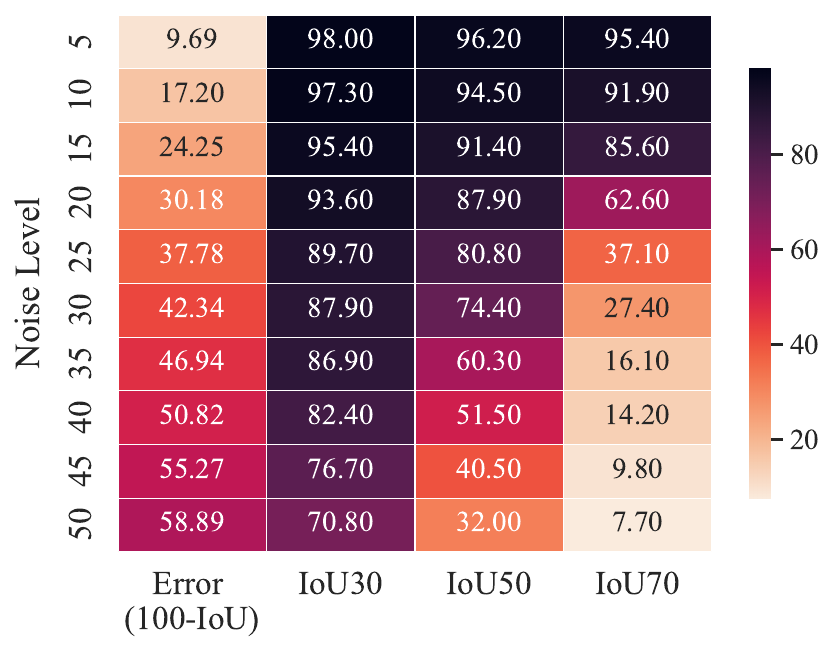}
        \end{minipage}} &

        \begin{minipage}[l]{0.03\textwidth}
            \vspace{-0.45cm}\subcaption{}\label{wacveval:fig:noisy-vs-gt-modelselec}
        \end{minipage} &
        \begin{minipage}[c]{0.3\textwidth}
            \centering
            \includegraphics[width=1\textwidth,trim=0 0 0 0.6cm,clip]{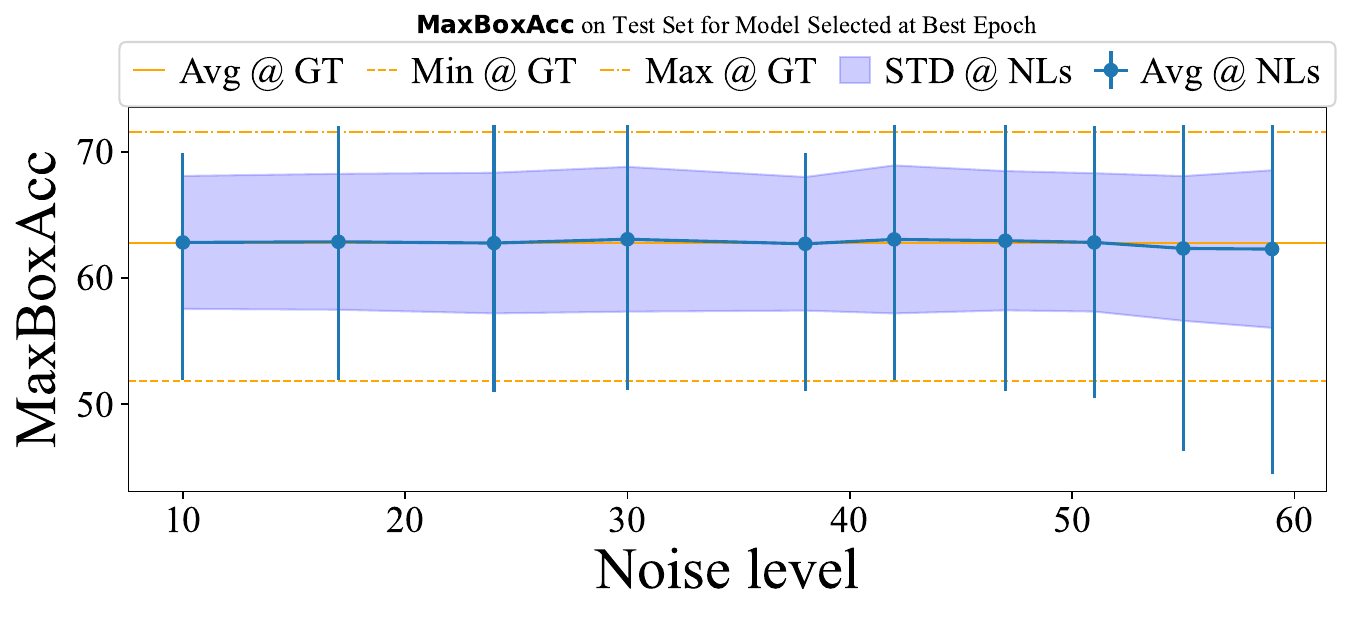}
        \end{minipage} 
        &
        \;
        \begin{minipage}[l]{0.03\textwidth}
            \vspace{-0.7cm}\subcaption{}\label{wacveval:fig:epochdiff-noisy-pseduo}
        \end{minipage} &
        \begin{minipage}[c]{0.3\textwidth}
            \includegraphics[width=0.98\textwidth,trim=0 0 0 2.14cm,clip]{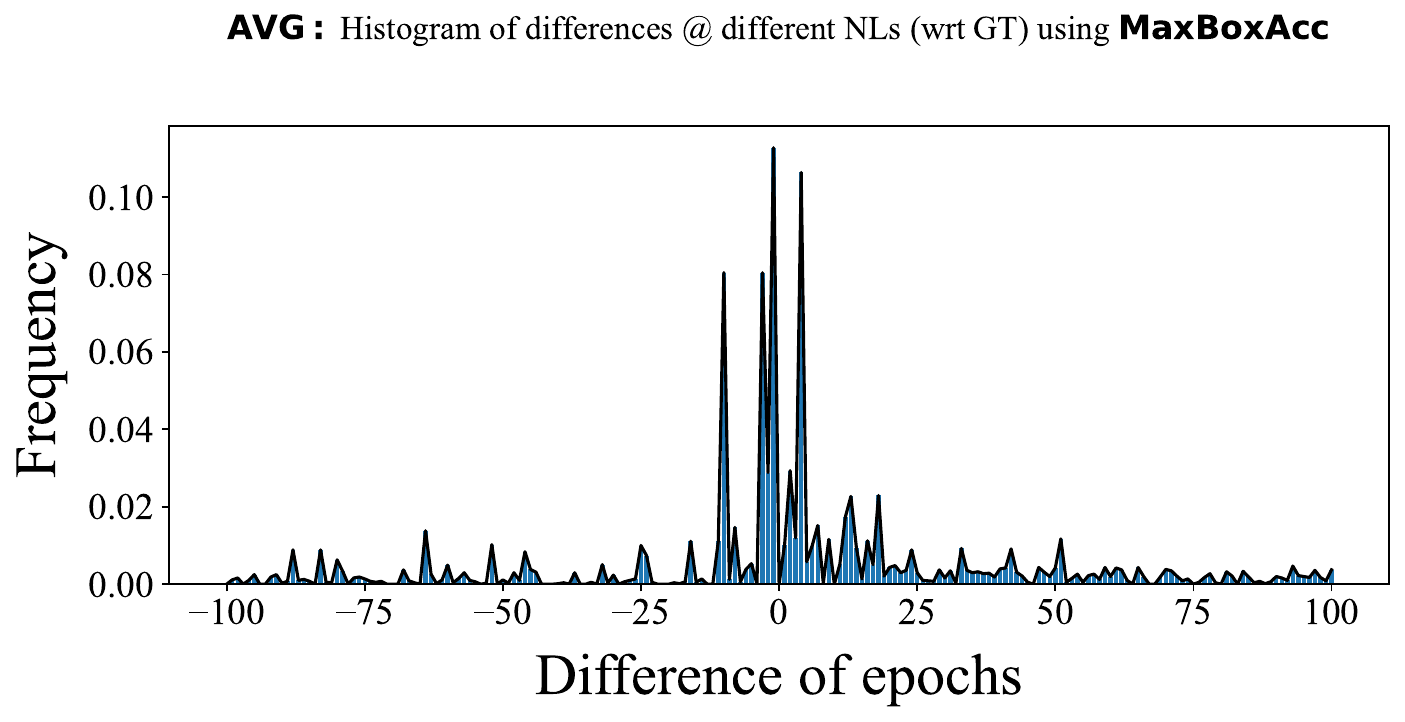}
        \end{minipage} \\
        
        \begin{minipage}[c]{0.015\textwidth}
        \end{minipage} &
        \begin{minipage}[c]{0.3\textwidth}
            \centering
        \end{minipage} 
        &
        
        \begin{minipage}[c]{0.03\textwidth}
            \vspace{-0.67cm}\subcaption{}\label{wacveval:fig:epoch-noisy-pseduo}
        \end{minipage} &
        \begin{minipage}[c]{0.3\textwidth}
            \centering
            \includegraphics[width=1\textwidth,trim=0 0 0 2.1cm,clip]{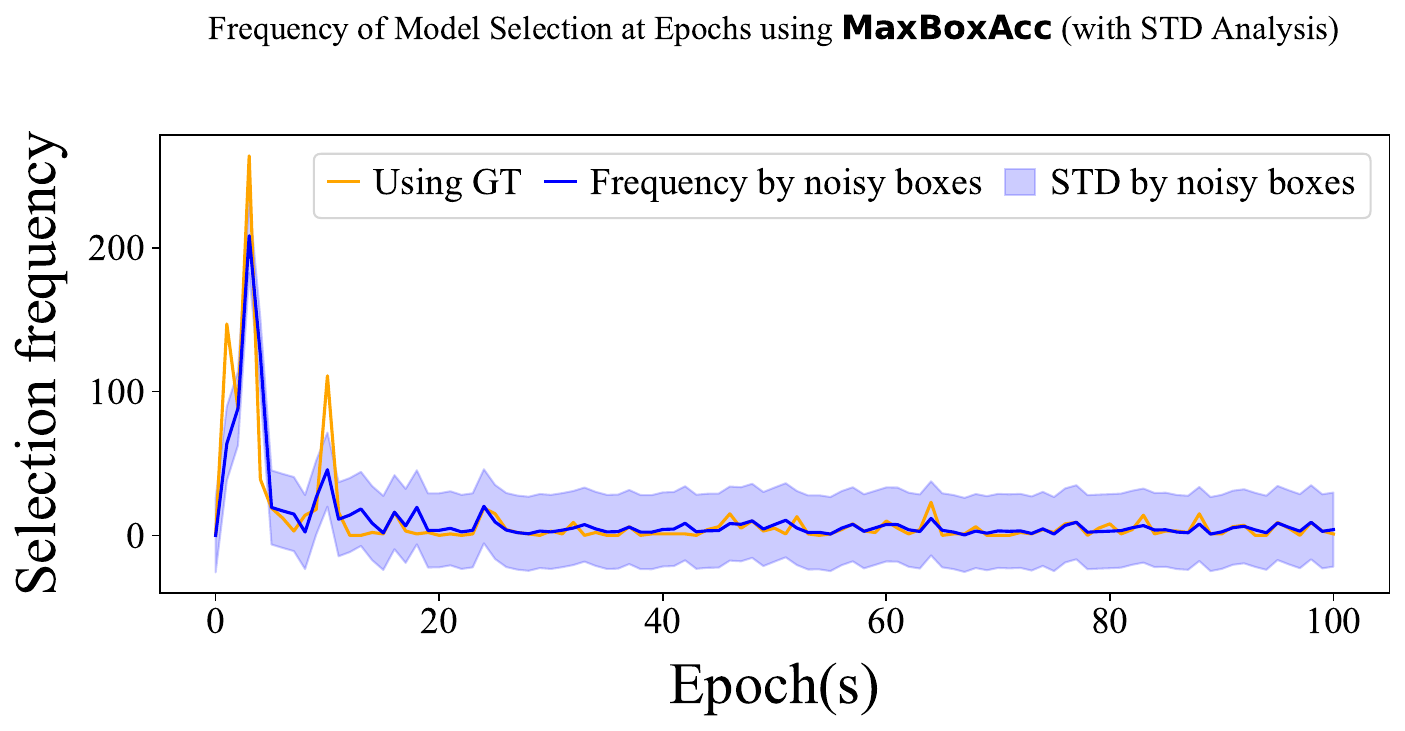}
        \end{minipage} &
        \;
        \begin{minipage}[c]{0.03\textwidth}
            \vspace{-0.67cm}\subcaption{}\label{wacveval:fig:pseduo-bbox}
        \end{minipage} &
        \begin{minipage}[c]{0.3\textwidth}
            \centering
            \hspace*{-0.28cm}
            \includegraphics[width=0.99\linewidth,trim=0 0cm 0 0cm,clip]{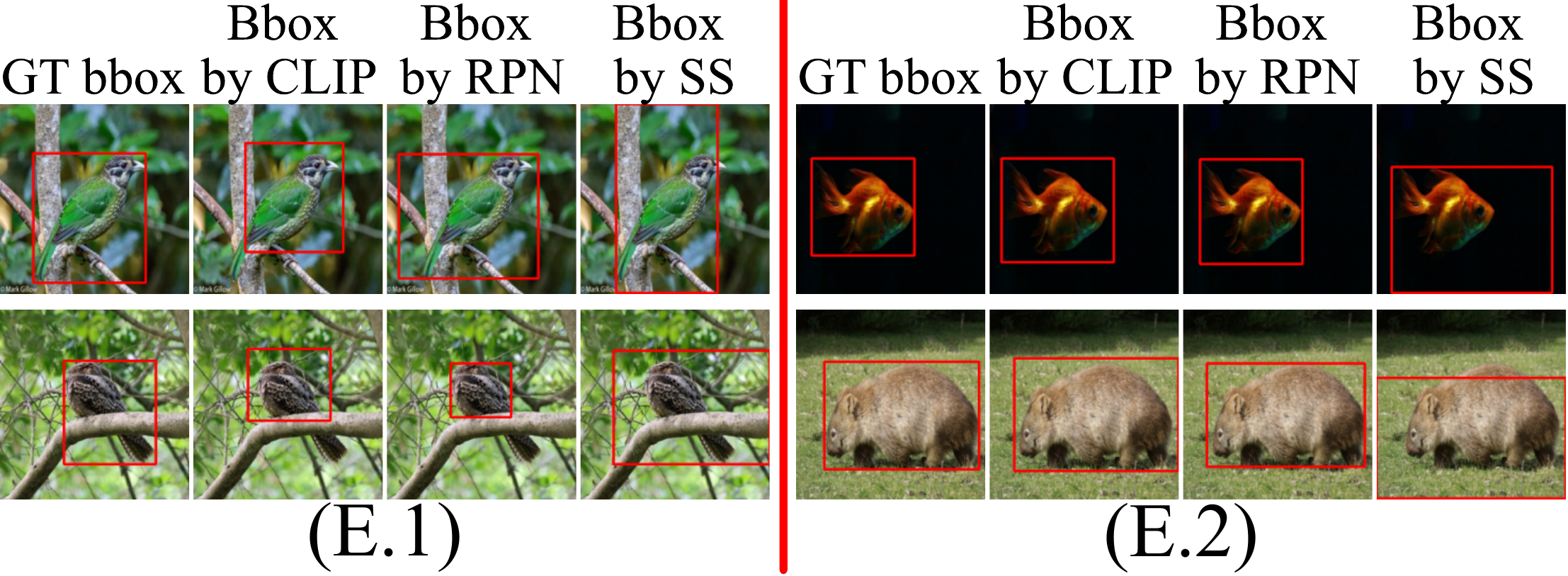}
        \end{minipage}
    \end{tabular}
    \vspace{-0.25cm}
    \caption{\textbf{(A)} The heatmap illustrates the inaccuracies in noisy bboxes at various noise levels generated by augmenting GT bboxes. 
    \textbf{(B)} The blue line represents the average \localization accuracy over various configurations on the  test set, incorporating maxima, minima, and standard deviation; the model is selected using a validation set comprising noisy bboxes with varying noise levels. In contrast, orange lines represent the average maximum and minimum performance of a model chosen during hyperparameter optimization over GT for configurations analogous to those employed with noisy labels.
    \textbf{(C)} Histogram illustrating the variance in model selection epochs when using bboxes at various noise levels w.r.t ground truth boxes for experiments with identical configurations, highlighting tendency around zero. 
    \textbf{(D)} Model selection epoch frequency using GT and noisy boxes across different configurations and noise levels. 
    \textbf{(E)} Illustration of pseudo-bboxes along with the GT over (E.1) \cubs and (E.2) \ilsvrc validation set.
    \vspace{-.38cm}}
    \label{wacveval:fig:main_figure}
\end{figure*}

\noindent\textbf{Implementation details.} 
For each dataset, we adopted a uniform batch size of 32, and resized each image to a size of $256\times256$ pixels and then employed random cropping to $224\times224$ pixels followed by random horizontal flipping. Experiments were performed on 50 training epochs for \cubs and 10 for \ilsvrc. Furthermore, each method was trained using the stochastic gradient descent optimizer while utilizing four shared hyperparameters for each method: learning rate, step size, weight decay, and gamma. For these shared hyperparameters, a total of 180 configurations were chosen. Additionally, for different methods, specific parameters were selected from their feasible range, and a cartesian product of these hyperparameters with the shared ones was computed to produce the final grid for hyperparameter\footnote{The comprehensive details on the hyperparameter space are included in the supplementary material.} optimization. This extensive set of configurations, substantially larger than the scope explored in prior studies, e.g. Choe et al. (2020), provides more fair comparisons among the different methods. Moreover, to study the impact of noisy bboxes on the model selection, we train CAM on \cubs until 100 epochs, exploring 1,080 configurations using common hyperparameters. 

\subsection{Results and Discussions}
\noindent\textbf{(1) Model selection with noisy ground truth bboxes.}\label{wacveval:sebsec:impact-noisy-boxes}
This section presents an initial results designed to validate the robustness of the proposed model selection techniques. We systematically perturbed the GT bbox, initially employed for model selection, to simulate the scenario for noisy or imprecise GT annotations. The primary aim is to evaluate the sensitivity of model selection to errors in the GT bbox. \textit{The objective is to determine whether growing degrees of GT bboxes noise affects model selection performance?} Positive results would substantiate the reliability of our model selection mechanism under real-world inaccuracies in bbox annotations. Our experiments encompass ten noise levels (see Fig.\ref{wacveval:fig:noislevel}), and involve 300 experiments using the CAM method with perturbed bboxes for validation set model selection. Our findings reveal that the WSOL method maintains its localization performance even with very noisy validation set bboxes (Fig.\ref{wacveval:fig:noisy-vs-gt-modelselec}). This suggests that model selection can be effectively performed with coarse localization cues. The selection frequency across different epochs using GT and noisy bboxes is remarkably similar (Fig.\ref{wacveval:fig:epoch-noisy-pseduo}). In Fig.\ref{wacveval:fig:epochdiff-noisy-pseduo} histogram comparing model selection epochs using GT and noisy bboxes indicates a central tendency around zero, showing that coarse GT bboxes from various sources enable reliable model selection while sustaining a high level of accuracy. \textit{More information regarding the generation of bboxes can be found in supplement materials.}

\noindent\textbf{(2) Generation of pseudo-GT bboxes.}
Three different off-the-shelf models, SS, RPN, and CLIP, are employed for generating pseudo-bboxes. Since, these methods generate a set class-agnostic pseudo-bboxes and \textit{pointing game} is employed for selecting discriminative boxes. This involves harvesting CAM from a pretrained classifier, pinpointing the peak activation, and evaluating its spatial overlap with annotated GT-bboxes. A successfull \localization of pointing game is counted if the peak activation in CAMs overlaps with annotated bboxes, using $\frac{Hits}{Hits+Misses}$. The pointing game accuracy of pretrained classifier's CAMs used to filter discriminative bboxes,
achieving 98.70\% accuracy on the \cubs validation split, 98.00\% on the test split, 88.05\% on the \ilsvrc validation split, and 89.03\% on the test split.

Details about generating pseduo-bboxes ${\hat{o}(x_i)}$ is presented in $\S$\ref{wacveval:sec:realisticprtocol}, and the \iou accuracy of the pseudo-bboxes w.r.t. GT bboxes on the validation set is reported in Tab.\ref{wacveval:tab:pseduobboxperf}. Despite the noise, these pseudo-boxes are expected to be useful for model selection and threshold estimation, as demonstrated in the previous section. Examples of pseudo-bboxes generated by RPN and SS are shown in Fig.\ref{wacveval:fig:pseduo-bbox}.

\renewcommand{\tabcolsep}{6pt}
\begin{table}[!t]
    \centering
    \resizebox{0.8\linewidth}{!}{%
    \begin{tabular}{|l|c|c|}
    \hline
         \multicolumn{1}{|c|}{\textbf{Annotator}} & \textbf{\cubs} & \textbf{\ilsvrc} \\\hline\hline
         \textbf{SS + Proposed bbox selection} & 39.98 & 45.07  \\\hline
         \textbf{RPN + Proposed bbox selection} & 71.23 & 61.08  \\\hline
         \textbf{CLIP + Proposed bbox selection} & 68.80 & 64.41 \\\hline
    \end{tabular}
    }
    \vspace{-0.2cm} 
    \caption{\iou performance of generated pseudo-bboxes w.r.t GT bboxes on the validation set.
    \vspace{-0.54cm}
    }
    \label{wacveval:tab:pseduobboxperf}
\end{table}

\noindent\textbf{(3) Pseudo-bboxes and $\tau$ estimation.}
We employed our proposed model selection protocol across different region proposal methods as outlined in~$\S$\ref{wacveval:sec:exp-meth}. During the training of each method, we employed a validation set for model selection with which encompassed image, class labels, pseudo-bboxes generated by different off-the-shelf methods. We compared the performance of the models trained with these pseudo-bboxes against those trained with ground truth bboxes. For every region proposal method, we selected a model using validation accuracy. Its performance was tested using a test set, and then the best-performing model was selected using the same type of supervision used to select the best on the validation set. The results from these experiments are reported in Tab.\ref{wacveval:tab:results_openimg_cub}, where WSOL models employing pseudo label bboxes for model selection achieved results comparable to those using GT bboxes. This suggests that reliable model selection is possible with coarse bboxes. 

\renewcommand{\tabcolsep}{1pt}

\begin{table}[!t]
\centering
\vspace{-0.1cm}
\resizebox{0.99\linewidth}{!}{%
\begin{tabular}{|l|c||c|c|c|c|c||c|c|c|c|c|}
\hline
\multicolumn{1}{|c}{}&\multicolumn{1}{c||}{}&  \multicolumn{5}{c||}{\textbf{\cubs (\maxboxacc)}} & \multicolumn{5}{c|}{\textbf{\ilsvrc (\maxboxacc)}} \\
 \cline{1-7} \cline{8-12}
\multicolumn{1}{|c|}{\textbf{Method}} & \textbf{Select} &  \textbf{\classification} & \textbf{\gt} & \textbf{\rpnlabels} & \textbf{\cliplabels} & \textbf{\sslabels} & \textbf{\classification} & \textbf{\gt} & \textbf{\rpnlabels} & \textbf{\cliplabels} & \textbf{\sslabels} \\
\hline\hline 
&    \btst & 30.82 & 91.56 & 92.25 & 92.26 & 91.76    &    47.55& 61.76 & 61.75 & 61.75 & 59.76 \\
\rowcolor{btestvalth} \cellcolor{white} & \btestvalth	 & -- & 90.26 & 92.00 & 92.16 & 26.44    &    -- & 61.76 & 61.15 & 61.38 & 49.27 \\
\rowcolor{bval} \cellcolor{white}& \bval	 & 30.82 & 91.56 & 92.25 & 92.26 & 80.16    &    47.55 & 61.76 & 61.75 & 61.75 & 59.76 \\
\rowcolor{bvalvalth} \cellcolor{white}\multirow{-4}{*}{\parbox{1.1cm}{\textbf{SCM}~\cite{bai2022weakly} {\small \emph{(eccv'22)} DeiT-S}}}     & \bvalvalth	 & -- & 90.26 & 92.00 & 92.16 & 22.02    &    -- & 61.76 & 60.21 & 61.38 & 49.27 \\
\cline{1-12} 
&    \btst & -- & -- & -- & -- & --    &    63.2 & 69.7 & 67.6 & 67.6 & 67.3 \\
\rowcolor{btestvalth} \cellcolor{white} & \btestvalth	 & -- & -- & -- & -- & --    &    -- & 69.3 & 68.2 & 68.2 & 60.8 \\
\rowcolor{bval} \cellcolor{white}& \bval	 & -- & -- & -- & -- & --    &    62.9 & 69.3 & 68.2 & 68.2 & 67.3 \\
\rowcolor{bvalvalth} \cellcolor{white}\multirow{-4}{*}{\parbox{1.42cm}{{\footnotesize\textbf{DiPS}}~\cite{Murtaza_2023_WACV} {\small \emph{(wacv'23)} DeiT}}}     & \bvalvalth	 & -- & -- & -- & -- & --    &    -- & 69.3 & 68.2 & 68.2 & 60.4 \\
\cline{1-12}
& \btst & 69.60 & 92.14 & 92.45 & 91.45 & 92.23    &    64.59 & 70.12 & 67.08 & 70.13 & 70.13 \\
\rowcolor{btestvalth} \cellcolor{white} & \btestvalth	 & -- & 91.66 & 92.26 & 91.00 & 41.62    &    -- & 69.46 & 64.78 & 69.05 & 57.06 \\
\rowcolor{bval} \cellcolor{white}& \bval	 & 70.24 & 91.33 & 91.75 & 89.97 & 91.33    &    66.17 & 70.12 & 67.08 & 70.13 & 70.13 \\
\rowcolor{bvalvalth} \cellcolor{white}\multirow{-4}{*}{\parbox{1.1cm}{\textbf{SAT}~\cite{wuspatial23} {\small \emph{(iccv'23)} DeiT-S}}}  & \bvalvalth	 & -- & 90.86 & 91.71 & 89.93 & 28.04 & -- & 69.46 & 64.78 & 69.05 & 57.06 \\
\hline
\end{tabular}
}
\vspace{-0.15cm}
\captionof{table}{
Test-set \maxboxacc of WSOL models with different selection criteria on \cubs and \ilsvrc. The \textit{select} column presents (i) BT and BV indicate model selection based on hyperparameter configurations using the test set and validation set, respectively; (ii) TT and VT indicate that the threshold \(\tau\) is selected using either the test set or validation set. For model selection on the validation set, we consider the GT as a reference, a selection based on the \classification performance and the three different pseudo-bboxes generation proposed in this work: RPN, CLIP and SS. Our results for models selected with pseudo-bboxes are comparable to those of GT.
\vspace{-0.5cm}
}
\label{wacveval:tab:results_openimg_cub}
\end{table}

Results also show that selecting models based on best validation (BV) across different hyperparameter configurations yields competitive results and is a more realistic approach when compared to the model selected using best test (BT) accuracy. Furthermore, when the best model is selected using pseudo-bboxes $\hat{o}(x_i)$ and the threshold $\tau$ is estimated using $\hat{o}(x_i)$ (BV-VT), we achieve accuracy comparable to methods that utilize manual bbox annotations for validation and use the BT for $\tau$ estimation (BT-TT). \textit{This indicates that model selection should be based on validation set performance across different experiments, avoiding overestimating model performance.} Additionally, if the foreground saliency mask predictor \cite{liu2010learning} is employed to fit oracle bboxes (converted into a binary map) over the ${\mathcal{D}_{val}}$, it achieves a \maxboxacc of 68.7\% and 94.0\% on the \ilsvrc and \cubs, respectively \cite{choe2020evaluating}. Performance is comparable to models trained in the WSOL setting, which utilize a much larger ${\mathcal{D}_{train}}$ and ${\mathcal{D}_{val}}$ for model selection. Our results also suggest that the pseudo-box annotation ${\mathcal{D}_{val}}$ should be used for model evaluation and $\tau$ estimation. 

\noindent\textbf{(4) Recommendation for realistic evaluation.} In line with our proposed evaluation protocol and empirical results, we recommend the following practices. Training and evaluation should be constrained to only utilizing class-level labels without resorting to manual bbox annotations for model selection and threshold estimation: 
\textbf{(i)} We recommend utilizing pseudo-bboxes generated by off-the-sheld methods for validation set. Our experiments show that these pseudo-bboxes can effectively provide the \localization annotation for model selection, achieving performance comparable to using GT annotations. \textbf{(ii)} Estimating thresholds using pseudo-bboxes from the validation set instead of the test set to produce bboxes from \localization maps. This approach addresses the unrealistic practice of thresholding.  
\textbf{(iii)} Our experiments suggest that model selection across various experiments, each employing different hyperparameters, should be based on the performance of pseudo-bboxes on the validation set. This approach mitigates bias in model performance evaluation. \textbf{(iv)} Be cautious when using thresholded-\iou metrics (\iou-30, \iou-50, \iou-70) and their variants, as these can mislead by considering uniform performance across instances above the set threshold. Our analysis reveals a non-linear relationship between thresholded-\iou and \iou (Tab.\ref{wacveval:tab:maxbox_vs_iou}). We recommend the use of non-thresholded-\iou for realistic evaluation.

\begin{table}[!t]
\centering
\vspace{-0.18cm}
\resizebox{0.9\linewidth}{!}{%
\begin{tabular}{|l||c|c|c|c|cc}
\hline
& \multicolumn{2}{c|}{\textbf{\cub}} & \multicolumn{2}{c|}{\textbf{\ilsvrc}} \\
\cline{1-7} \cline{6-7}
\multirow{2}{*}{\textbf{Method}} & \textbf{\maxboxacc} & \multirow{2}{*}{\quad\;\;\;\textbf{\iou}\quad\;\;\;} & \textbf{\maxboxacc} & \multirow{2}{*}{\quad\;\;\;\textbf{\iou}\quad\;\;\;} \\
& \textbf{(\iou-50)} & & \textbf{(\iou-50)} & \\
\hline\hline
CAM~\cite{zhou2016learning} & 70.40 & 56.71 & 64.06 & 57.89 \\
HaS~\cite{singh2017hide} & 75.85 & 59.81 & 63.77 & 58.49 \\
ACoL~\cite{zhang2018adversarial} & 74.64 & 58.29 & 62.93 & 56.39 \\
ADL~\cite{choe2019attention} & 76.63 & 60.12 & 65.11 & 58.46 \\
NL-CCAM~\cite{yang2020combinational} & 65.58 & 54.76 & 60.63 & 54.98 \\
TS-CAM~\cite{gao2021ts} & 90.19 & 69.78 & 66.75 & 59.54 \\
SCM~\cite{bai2022weakly} & 91.56 & 70.27 & 61.76 & 54.55 \\
SAT~\cite{wuspatial23}  & 92.14 & 73.67 & 70.12 & 62.80 \\
\hline
\end{tabular}
}
\vspace{-0.15cm}
\captionof{table}{Analysis of \maxboxacc(\iou-50) versus \iou using oracle bboxes for model selection. Analysis using pseudo-bboxes is presented in \textit{supplementary material} exhibiting same behaviour.\vspace{-0.55cm}}
\label{wacveval:tab:maxbox_vs_iou}
\end{table}

\section{Conclusion}
\label{wacveval:sec:conclusion}
The currently-used protocol~\cite{choe2020evaluating} for evaluation of WSOL methods has driven significant progress in the field. However, its reliance on manually annotated bboxes during validation for model selection, and annotated test set for threshold estimation leads to an overestimation of localization performance on the test set. In this work, we propose generating pseudo-bboxes for the validation set using off-the-shelf pretrained region proposal models for both model selection and threshold estimation. This threshold is then employed to produce bboxes from \localization maps on the test set. Our approach was evaluated with different WSOL methods and shows that employing pseudo-bboxes for model evaluation achieves localization performance comparable to models selected using GT bboxes where the threshold is estimated on the test set. Our approach is therefore a promising and more realistic alternative than using a GT annotated held-out dataset. Despite the promising results with our protocol, performing model selection in WSOL without \localization annotation remains an open issue. 

\noindent\textbf{Supplementary materials} include a description of evaluated methods, steps to generate noisy GT bboxes, the performance of pseudo-bboxes at different selection steps, the impact of thresholding on results, and results on medical datasets. Results obtained using addition methods, specifically F-CAM~\cite{belharbi2022fcam} and DiPS~\cite{murtaza2023dips}, are available: \href{https://github.com/shakeebmurtaza/wsol_model_selection}{github.com/shakeebmurtaza/wsol\_model\_selection}.

\noindent\textbf{Acknowledgements.} This research was supported by the Natural Sciences and Engineering Research Council of Canada. We also thank the Digital Research Alliance of Canada for the use of their computing resources. 

%% file: sections/2_supp_content.tex
\appendix
\newcommand{\hbAppendixPrefix}{S}

\renewcommand{\thefigure}{\hbAppendixPrefix\arabic{figure}}
\setcounter{figure}{0}
\renewcommand{\thetable}{\hbAppendixPrefix\arabic{table}} 
\setcounter{table}{0}
\renewcommand{\theequation}{\hbAppendixPrefix\arabic{equation}} 
\setcounter{equation}{0}

\noindent This supplementary material contains the following content:
\begin{enumerate}[label=\textbf{\Alph*.},itemsep=0pt, parsep=0pt]
    \item \textbf{Review of evaluated methods and their search spaces.}
        In this section, we provided a detailed description of methods employed to evaluate the efficacy of our proposed protocol for model selection. Additionally, it outlines various hyperparameters for each method and their respective feasible range.
    \item \textbf{Generation of noisy ground truth bboxes.} This section delineates the methodology for generating noisy ground truth (GT) bounding boxes (bboxes) to analyze their impact on model selection.
    \item \textbf{Pseudo-bboxes performance across different selection steps.} This section presents the performance of pseudo-bboxes at different selection steps of our proposed method. 
    \item \textbf{Localization evaluation in WSOL (thresholded-\iou vs \iou).} This section describes an ongoing challenge in WSOL where discrepancies in localization performance exist between the commonly used \maxboxacc and the \iou metric.
    \item \textbf{Localization Evaluation with different thresholds.} This section presents the results at different thresholds along with their average.
    \item  \textbf{Experiments with medical datasets.} To show the generalizability of our proposed evaluation protocol across various domains, we employ medical datasets.
\end{enumerate}

\section{Review of Evaluated Methods and their Search Spaces}
\subsection{Evaluated Methods}
To assess the efficacy of the proposed protocol, we employ eight methods published in top-tier venues from 2016 to 2023. These methods are presented chronologically within this section, providing a comprehensive overview.

\noindent\textbf{Class activation mapping (CAM)~\cite{zhou2016learning}} is able to extract an activation map for a particular class using a pre-trained CNN-classifier with global average pooling. It generates the final map by aggregating different activation maps from the penultimate convolution layer based on the contribution towards each class by using weights from the last fully connected layer. 

\noindent\textbf{Hide-and-seek (HaS)~\cite{singh2017hide}} force the network to look beyond discriminative regions of a particular object by augmenting the input image. It hides patches of input image during training by employing two hyper-parameters; drop rate and grid size. 

\noindent\textbf{Adversarial complementary learning (ACoL)~\cite{zhang2018adversarial}} employees an architecture with two parallel classifier heads that tires to find complementary regions by adversarially erasing high-scoring activations. 

\noindent\textbf{Attention-based dropout layer (ADL)~\cite{choe2019attention}} works similarly as ACoL by erasing high-score activation to force by employing drop masks generated without second classifiers head.

\noindent\textbf{Non-local combinational class activation maps (NL-CCAM)~\cite{yang2020combinational}.} In this paper, the author argues that employing the activation map of the class with the highest classifier's score may only highlight discriminative regions and for certain images it tends to focus on background regions. To address this limitation, the author proposes to combine activation maps of different classes. This combination is based on the respective probability score of each, encompassing a spectrum from the highest to the lowest.

\noindent\textbf{Token semantic coupled attention map (TS-CAM)~\cite{gao2021ts}.} TS-CAM is a cascaded ViT-CNN architecture that proposes to redistribute class information to patch tokens. This is achieved by implementing a CNN-based classification (\classification) head atop the patch tokens, thereby rendering the \cls token class-agnostic.  
Therefore, these \cls tokens are combined with the activation map extracted from the last convolutional layer to produce an activation map highlighting different object parts of a particular class.

\noindent\textbf{Spatial calibration module (SCM)~\cite{bai2022weakly}.} This paper introduced an SCM module atop the transformer features to align the boundaries of the generated map with the object boundaries by avoiding partial activation in different areas of the activation map.  This module integrates semantic similarities presented in patch tokens and their spatial relationships into a unified model. SCM effectively recalibrates the transformer’s attention and semantic representations to mitigate the background noise and sharpen object boundaries. 

\noindent\textbf{Task-specific spatial-aware token (SAT).}~\cite{wuspatial23} This paper introduces a spatial-aware token (SAT) into the transformer's input space. Like \cls token that is able to accumulate information for \classification tasks, it is incorporated to aggregate the global representation of the object of interest. Furthermore, the SAT is a passed-to spatial-query attention module that treats the SAT as a query to calculate similarity with different patches and produces probabilities for foreground object for producing accurate localization maps. 

\subsection{Hyperparamter Search Space}
To fairly compare different WSOL methods, we took steps to minimize human biases during training. This includes employing pseudo-bboxes for the evaluation and sampling hyperparameter values from the feasible range, except for annotations on the test set, which were used only to assess the trained model's performance. 

Each method was trained using four shared hyperparameters, while the additional hyperparameters were specific to each model. We sampled the values for these hyperparameters from their feasible range. A cartesian product of these and shared hyperparameters was computed to create the final grid of hyperparameters for training each model. A detailed summary of the hyperparameters employed to train different WSOL models is presented in Tab.\ref{wacveval:tab:supp-params}.

\begin{table}[!h]
    \footnotesize
    \centering
    \resizebox{1\linewidth}{!}{%
    \begin{tabular}{|c|l|c|c|}
        \hline
        \textbf{Method} & \multicolumn{1}{c|}{\textbf{Hyperparameter}} & \multicolumn{1}{c}{\parbox[c][0.8cm][c]{1.5cm}{\centering \textbf{Sampling}\\\textbf{Distribution}}} & \multicolumn{1}{|c|}{\textbf{Range}} \\
        \hline\hline
        \multirow{3}{*}{Common HPs} & LR, WD, Gamma & LogUniform & $[10^{-5},10^0]$ \\
        \cline{2-4}
        & Step Size & Uniform & \begin{tabular}[c]{@{}l@{}}\cubs: $[ 5-45]$\\ \ilsvrc: $[ 2-9]$\end{tabular} \\
        \hline
        \multirow{2}{*}{\begin{tabular}[c]{@{}c@{}}CAM~\cite{zhou2016learning}, TS-CAM~\cite{gao2021ts} \\ SCM~\cite{bai2022weakly}, NL-CCAM~\cite{yang2020combinational}\end{tabular}} & \multirow{2}{*}{Common HPs} & \multirow{2}{*}{-} & \multirow{2}{*}{-} \\
        & & & \\
        \hline
        HaS~\cite{singh2017hide} & Drop Rate, Drop Area & Uniform & $[0,1]$ \\
        \hline
        ACoL~\cite{zhang2018adversarial}  & Erasing Threshold & Uniform & $[0,1]$ \\
        \hline
        ADL~\cite{choe2019attention} & Drop Rate, Erasing Threshold & Uniform & $[0,1]$ \\
        \hline
        SAT~\cite{wuspatial23} & Area Threshold & Uniform & $[0,1]$ \\
        \hline
    \end{tabular}
    }
    \caption{Hyperparameter search space for different methods}\label{wacveval:tab:supp-params}
\end{table}

\begin{table}[!b]
\centering
\resizebox{1\linewidth}{!}{%
\begin{tabular}{|l|ccc|ccc|}
\hline
\multicolumn{1}{|c|}{\multirow{2}{*}{\textbf{Method}}} &
  \multicolumn{3}{c|}{\textbf{\cubs (\iou)}} &
  \multicolumn{3}{c|}{\textbf{\ilsvrc (\iou)}} \\ \cline{2-7}\noalign{\vskip 2pt}\cline{2-7}
\multicolumn{1}{|c|}{} &
  \multicolumn{1}{c|}{\textbf{SS}} &
  \multicolumn{1}{c|}{\textbf{RPN}} &
  \textbf{CLIP} &
  \multicolumn{1}{c|}{\textbf{SS}} &
  \multicolumn{1}{c|}{\textbf{RPN}} &
  \textbf{CLIP} \\ \hline
\multicolumn{1}{|l|}{Mean IoU (PG$^{*}$)} &
  \multicolumn{1}{c|}{32.21} &
  \multicolumn{1}{c|}{28.71} &
  -- &
  \multicolumn{1}{c|}{23.38} &
  \multicolumn{1}{c|}{27.57} &
  -- \\ \hline
\multicolumn{1}{|l|}{Mean IoU (PG$^{*}$, 20\% Filtered)} &
  \multicolumn{1}{c|}{33.69} &
  \multicolumn{1}{c|}{37.89} &
  -- &
  \multicolumn{1}{c|}{34.99} &
  \multicolumn{1}{c|}{37.17} &
  -- \\ \hline
\multicolumn{1}{|l|}{\parbox{4cm}{Mean IoU (Top Box, \\ 20\% by Objectness Score)}} &
  \multicolumn{1}{c|}{39.90} &
  \multicolumn{1}{c|}{69.80} &
  -- &
  \multicolumn{1}{c|}{44.90} &
  \multicolumn{1}{c|}{54.02} &
  -- \\ \hline
\multicolumn{1}{|l|}{\parbox{4cm}{Mean IoU (Top Box,\\ 20\%, PG$^{*}$+Scoring)}} &
  \multicolumn{1}{c|}{39.98} &
  \multicolumn{1}{c|}{71.23} &
  -- &
  \multicolumn{1}{c|}{45.07} &
  \multicolumn{1}{c|}{61.08} &
  -- \\ \hline
\multicolumn{1}{|l|}{\parbox{4cm}{Upper Bound (Select by \\IoU with GT)}} &
  \multicolumn{1}{c|}{64.07} &
  \multicolumn{1}{c|}{83.66} &
  -- &
  \multicolumn{1}{c|}{65.46} &
  \multicolumn{1}{c|}{84.42} &
  -- \\ \hline
\multicolumn{1}{|l|}{IoU from Otsu} &
  \multicolumn{1}{c|}{--} &
  \multicolumn{1}{c|}{--} &
  68.80 &
  \multicolumn{1}{c|}{–} &
  \multicolumn{1}{c|}{–} &
  64.41 \\ \hline
\multicolumn{1}{|l|}{IoU using 1K Threshold} &
  \multicolumn{1}{c|}{--} &
  \multicolumn{1}{c|}{--} &
  69.56 &
  \multicolumn{1}{c|}{–} &
  \multicolumn{1}{c|}{–} &
  65.78 \\ \hline
  \multicolumn{1}{l}{\hspace*{-0.3cm} \footnotesize $^{*}$PG: Pointing Game}
 
\end{tabular}
}
\captionof{table}{Performance of pseudo-bboxes obtained using different off-the-shelf region proposal methods across multiple refinement stages over the validation set. The table highlights the incremental improvement in \iou through various selection steps.
}
\label{wacveval:tab:supp-pseduobboxperf}
\end{table}

\begin{table*}[!t]
\centering
\resizebox{1\linewidth}{!}{%
\begin{tabular}{|l|c|c|c|c|c|c|c|c|c|c|c|c|c|c|c|c|c|c|c|c|c|c|c|c|c|c|c|}
\hline
\multirow{2}{*}{\textbf{Method}} & \multicolumn{4}{c|}{\textbf{\multirow{2}{*}{\backbonebold}}}& \multicolumn{6}{c|}{\textbf{\cub (\maxboxacc)}} & \multicolumn{6}{c|}{\textbf{\cub (\iou)}} & \multicolumn{6}{c|}{\textbf{\ilsvrc (\maxboxacc)}} & \multicolumn{5}{c|}{\textbf{\ilsvrc (\iou)}} \\
 \cline{6-11} \cline{12-28}
\textbf{} & \multicolumn{4}{c|}{\textbf{}} &  \textbf{\classification} & \textbf{\gt} & \textbf{\rpnlabels} & \textbf{\cliplabels} & \multicolumn{2}{c|}{\textbf{\sslabels} } & \textbf{\classification} & \textbf{\gt} & \textbf{\rpnlabels} & \textbf{\cliplabels} & \multicolumn{2}{c|}{\textbf{\sslabels}} &  \textbf{\classification} & \textbf{\gt} & \textbf{\rpnlabels} & \textbf{\cliplabels} & \multicolumn{2}{c|}{\textbf{\sslabels}} & \textbf{\classification} & \textbf{\gt} & \textbf{\rpnlabels} & \textbf{\cliplabels} & \textbf{\sslabels} \\
\hline\hline
CAM~\cite{zhou2016learning}             {\small \emph{(cvpr,2016)}} & \multicolumn{4}{c|}{ResNet50} & 66.98 & 70.40 & 71.10 & 70.62 & \multicolumn{2}{c|}{69.89} & 55.53 & 56.71 & 56.88 & 56.65 & \multicolumn{2}{c|}{56.76} &  61.48 & 64.06 & 63.60 & 63.90 & \multicolumn{2}{c|}{63.60} & 56.17 & 57.89 & 57.42 & 57.68 & 57.42 \\
HaS~\cite{singh2017hide}                {\small \emph{(iccv,2017)}} & \multicolumn{4}{c|}{ResNet50} & 67.62 & 75.85 & 75.85 & 75.85 & \multicolumn{2}{c|}{74.73} & 57.01 & 59.81 & 59.81 & 59.81 & \multicolumn{2}{c|}{59.39} & 61.69 & 63.77 & 63.94 & 63.77 & \multicolumn{2}{c|}{63.30} & 56.81 & 58.49 & 58.39 & 58.49 & 57.32 \\
ACoL~\cite{zhang2018adversarial}        {\small \emph{(cvpr,2018)}} & \multicolumn{4}{c|}{ResNet50} & 66.62 & 74.64 & 74.64 & 75.37 & \multicolumn{2}{c|}{74.14} & 55.32 & 58.29 & 58.29 & 58.55 & \multicolumn{2}{c|}{58.13} & 61.98 & 62.93 & 62.75 & 63.45 & \multicolumn{2}{c|}{62.92} & 55.62 & 56.39 & 56.32 & 56.94 & 56.39 \\
ADL~\cite{choe2019attention}            {\small \emph{(cvpr,2019)}} & \multicolumn{4}{c|}{ResNet50} & 67.82 & 76.63 & 76.63 & 76.06 & \multicolumn{2}{c|}{74.99} & 55.64 & 59.12 & 59.12 & 58.93 & \multicolumn{2}{c|}{58.32} & 62.81 & 65.11 & 65.97 & 65.11 & \multicolumn{2}{c|}{65.19} & 56.39 & 58.46 & 58.37 & 58.55 & 58.54\\
NL-CCAM~\cite{yang2020combinational} {\small \emph{(wacv,2020)}} & \multicolumn{4}{c|}{VGG-GAP} & 64.15 & 65.58 & 65.58 & 65.22 & \multicolumn{2}{c|}{45.97} & 54.11 & 54.76 & 54.76 & 54.59 & \multicolumn{2}{c|}{47.88} & 58.42 & 60.63 & 60.63 & 60.63 & \multicolumn{2}{c|}{52.72} & 51.59 & 54.98 & 54.98 & 54.98 & 49.44 \\
TS-CAM~\cite{gao2021ts} {\small \emph{(iccv,2021)}} & \multicolumn{4}{c|}{DeiT-S} & 88.36 & 90.19 & 90.35 & 89.52 & \multicolumn{2}{c|}{88.71} & 69.14 & 69.78 & 69.83 & 69.95 & \multicolumn{2}{c|}{68.36} & 56.40 & 66.75 & 66.75 & 66.75 & \multicolumn{2}{c|}{66.17} & 53.67 & 59.54 & 59.54 & 59.54 & 59.00 \\
SCM~\cite{bai2022weakly} {\small \emph{(eccv,2022)}} & \multicolumn{4}{c|}{DeiT-S} & 88.47 & 91.56 & 92.25 & 92.26 & \multicolumn{2}{c|}{91.76} & 68.64 & 70.27 & 70.89 & 70.93 & \multicolumn{2}{c|}{70.34} & 57.92 & 61.76 & 61.75 & 61.75 & \multicolumn{2}{c|}{59.76} & 52.13 & 54.55 & 54.56 & 54.56 & 53.47 \\
F-CAM~\cite{belharbi2022fcam} {\small \emph{(wacv,2022)}} & \multicolumn{4}{c|}{ResNet50} & 24.95 & 89.83 & 89.23 & 89.81 & \multicolumn{2}{c|}{88.26} & 37.72 & 68.79 & 68.12 & 68.62 & \multicolumn{2}{c|}{68.30} & -- & -- & -- & -- & \multicolumn{2}{c|}{--} & -- & -- & -- & -- & -- \\
SAT~\cite{wuspatial23} {\small \emph{(iccv,2023)}} & \multicolumn{4}{c|}{DeiT-S} & 79.70 & 92.14 & 92.45 & 91.45 & \multicolumn{2}{c|}{92.23} & 63.13 & 73.67 & 73.59 & 72.92 & \multicolumn{2}{c|}{73.61} & 64.94 & 70.12 & 67.08 & 70.13 & \multicolumn{2}{c|}{70.13} & 56.09 & 62.80 & 58.55 & 62.80 & 62.80  \\
\hline
\end{tabular}
}
\captionof{table}{
Comparative Analysis of \maxboxacc(\iou-50) versus \iou on \cubs and \ilsvrc with different model selection methods. }
\label{wacveval:tab:all_maxbox_vs_iou}
\end{table*}

\section{Generation of Noisy Ground Truth Bboxes}
In the main paper, we introduce a validation protocol designed to evaluate the robustness of the proposed model selection techniques in the presence of noisy GT bboxes. This protocol systematically perturbs the GT bboxes, initially used for model selection, to emulate conditions of noisy or imprecise GT annotations. To produce noisy GT bboxes, a sequence of random transformations is applied to the GT bboxes, creating varying noise levels. For each transformation, a total of ten unique noise levels are defined. These levels signify the maximum likelihood of deformation at each noise level. This likelihood is derived by sampling the deformation value using a uniform distribution that varies from 5 to 50, with intervals set at 5. To generate nosy bboxes we, first apply, scaling transformation to the GT bboxes between -50\% and +50\% with a maximum likelihood of a particular noise level. Following the scaling, we apply shift transformation to the scaled bbox by choosing a random shift length, where the shift length is set between 0\% and the maximum size percentage corresponding to a particular noise level. Finally, we modify the aspect ratio of bbox based on a probability factor `p' which indicates the likelihood, representing a specific noise level.

\section{Pseudo-bboxes Performance Across Different Selection Steps}
Different off-the-shelf models, SS, RPN, and CLIP, are employed to produce pseudo-bboxes. These methods generate a set of class-agnostic pseudo-bboxes. To select discriminative boxes from a pool of object proposals, the pointing game analysis was employed~\cite{zhang2018top}. This involves harvesting CAM from a pre-trained classifier and pinpointing the peak activation that is used to select discriminative boxes. Despite the initial filtering of bboxes via the pointing game, a substantial number of bboxes remained. To address this, we employ a sequential refinement process, in which we initially filter the top 20\% based on objectness or classifier score for boxes obtained from RPN and SS, respectively. Subsequently, the pointing game was employed to refine this selection, followed by selection of top-performing boxes based on score. In the case of CLIP, we utilized Otsu's thresholding method to identify binary maps, upon which bboxes were delineated around the largest connected areas. A comprehensive description of the proposed method for generating pseudo-bboxes is provided in the main paper.

The performance of pseudo-bboxes at different selection steps is presented in Tab.\ref{wacveval:tab:supp-pseduobboxperf}. This table shows that as we select relevant bboxes generated by SS or RPN at each stage, we progressively choose better-performing bboxes, resulting in reliable performance relative to the upper bound performance when using GT bboxes to select top-performing bbox. Initially, pseudo-bboxes generated by SS and RPN are filtered based on objectness or classifier scores, followed by a pointing game analysis for further refinement. The results indicate a significant improvement in the mean intersection over union (\iou) across these selection stages. For example, after the initial selection and filtering of the top 20\% based on objectness scores, the \iou increases substantially, demonstrating the efficacy of our proposed method. In contrast, CLIP generates activation maps that highlight particular objects. Otsu's thresholding method is employed to convert these maps into binary images, enabling the delineation of bboxes around the largest connected areas. Despite the single-stage selection process for CLIP-generated maps, the resulting bounding boxes achieve competitive performance.

\section{Localization Evaluation in WSOL (Thresholded-\iou vs \iou)}
\label{wacveval:sec:iou}
So far, we have reported the localization performance using \maxboxacc metric~\cite{choe2020evaluating}. It is also known as as \textit{GT-known localization} metric. It scores one point when the \iou between the GT bbox and the predicted box is above ${50\%}$, otherwise, it scores 0. It is referred to \iou5-0 as well. It is a well established and commonly used metric in WSOL. In addition to \iou-50, we report the \iou in Tab.\ref{wacveval:tab:maxbox_vs_iou} of the main paper. These results show that model selection using \classification accuracy still lead to poor \iou. However, selection using our proposed pseudo-bboxes yields competitive \iou compared to when using oracle bboxes.

The comparison based on \iou (Tab.\ref{wacveval:tab:maxbox_vs_iou} of the main paper) lead us to an interesting result presented in Tab.\ref{wacveval:tab:all_maxbox_vs_iou}. This table shows that \maxboxacc, using the commonly oracle bboxes, gives largely higher localization scores compared to the exact localization accuracy reported by \iou. For instance, SAT method~\cite{wuspatial23} scores ${92.14\%}$ in \maxboxacc, while it only scores ${73.12\%}$ over \cubs dataset. When considering only \maxboxacc, the results give the impression that \cubs dataset is saturated, especially when the same authors~\cite{wuspatial23} have reported a \maxboxacc of ${98.45\%}$. However, when inspecting \iou metric, localization is still low at ${73.12\%}$. 

In addition, since \maxboxacc is based on thresholding, extreme localization scores can hit the same scoring point. For instance, a prediction with \iou${=50.1\%}$ scores the same point as when the prediction is \iou${=99.99\%}$. However, both \iou${=49.9\%}$, \iou${=1\%}$ scores 0 point in \maxboxacc. This makes localization evaluation less efficient.

Despite its common usage in the literature, the aforementioned limitations of \maxboxacc suggest that reporting \iou along with \maxboxacc could be beneficial in better assessing localization performance of different methods in WSOL.

Since \cubs dataset is relatively easier than \ilsvrc dataset, the latter paints a realistic evaluation of the progress that has been done in WSOL. Since the work of Zhou et al.~\cite{zhou2016learning} in 2016 up to now, only ${\approx6\%}$, and ${\approx4.9\%}$ of improvement has been done in term of \maxboxacc and \iou, respectively. This suggests that a lot of work is still needs to be done to furthermore improve WSOL methods.

\section{Localization Evaluation with different thresholds}
\label{wacveval:sec:theval}
We have previously evaluated localization performance using the \maxboxacc metric, which assigns a score of one when the \iou between the GT bbox and the predicted bounding box exceeds 50\%, and a score of zero otherwise, as well as using raw \iou without any threshold. In addition, we extend the evaluation of \maxboxacc by examining performance trends at multiple \iou thresholds, namely \iouthirty, \ioufifty, and \iouseventy, along with their average, termed \newmaxboxacc, a metric commonly used in the weakly supervised object localization (WSOL) literature \cite{choe2020evaluating}. The performance results for these thresholded \iou metrics and their average are detailed in Tab.\ref{wacveval:tab:supp-results_cub_detialed}\&\ref{wacveval:tab:supp-results_ilsvrc_detialed}. In contrast to raw \iou, the thresholded \iou metrics demonstrate consistent trends when compared to \maxboxacc, which are sufficient to offer meaningful insights into the results.

\setlength{\tabcolsep}{1pt}
\begin{table*}[!t]
\centering
\resizebox{1\linewidth}{!}{%
\begin{tabular}{|l||c||c|c|c|c||c|c|c|c||c|c|c|c||c|c|c|c||c|c|c|c|c|}
\hline
						& & \multicolumn{4}{c|}{\textbf{CL}}  & \multicolumn{4}{c|}{\textbf{GT}}  & \multicolumn{4}{c|}{\textbf{RPN}}  & \multicolumn{4}{c|}{\textbf{CLIP}}  & \multicolumn{4}{c|}{\textbf{SS}} \\\hline

\textbf{Method} & \textbf{Select} & \iouthirty  & \ioufifty  & \iouseventy  &  {\footnotesize\newmaxboxaccs}  & \iouthirty  & \ioufifty  & \iouseventy  &  {\footnotesize\newmaxboxaccs}  & \iouthirty  & \ioufifty  & \iouseventy  &  {\footnotesize\newmaxboxaccs}  & \iouthirty  & \ioufifty  & \iouseventy  &  {\footnotesize\newmaxboxaccs}  & \iouthirty  & \ioufifty  & \iouseventy  &  {\footnotesize\newmaxboxaccs} \\\hline

& \btst & 73.76 & 32.99 & 6.161 & 37.63 &   93.95 & 70.40 & 20.05 & 61.46 &   94.30 & 71.10 & 19.74 & 61.71 &   94.33 & 70.62 & 19.46 & 61.47 &   94.06 & 69.89 & 20.38 & 61.44      \\
\rowcolor{btestvalth} \cellcolor{white} & \btestvalth & -- & -- & -- & -- &   93.13 & 69.50 & 19.07 & 60.56 &   94.25 & 70.43 & 19.22 & 61.30 &   93.73 & 69.88 & 18.46 & 60.69 &   70.05 & 27.97 & 12.80 & 36.94      \\
\rowcolor{bval} \cellcolor{white} & \bval & 69.72 & 29.37 & 5.143 & 34.74 &   93.95 & 70.40 & 20.05 & 61.46 &   93.95 & 70.40 & 20.05 & 61.46 &   94.70 & 69.76 & 18.46 & 60.97 &   90.05 & 61.20 & 16.44 & 55.89      \\
\rowcolor{bvalvalth} \cellcolor{white}\multirow{-4}{*}{\parbox{1cm}{\textbf{CAM}~\cite{zhou2016learning} {\small \emph{(cvpr'16)} ResNet50}}} & \bvalvalth & -- & -- & -- & -- &   93.61 & 69.20 & 19.33 & 60.71 &   93.61 & 68.93 & 19.03 & 60.52 &   94.70 & 67.98 & 16.58 & 59.75 &   58.66 & 22.79 & 15.84 & 32.43      \\
\hline

& \btst & 71.48 & 37.00 & 11.25 & 39.91 &   94.59 & 75.85 & 28.49 & 66.31 &   94.59 & 75.85 & 28.49 & 66.31 &   94.59 & 75.85 & 28.49 & 66.31 &   94.64 & 74.73 & 27.42 & 65.59      \\
\rowcolor{btestvalth} \cellcolor{white} & \btestvalth & -- & -- & -- & -- &   92.33 & 74.68 & 29.15 & 65.38 &   94.44 & 75.69 & 28.27 & 66.13 &   94.14 & 75.69 & 27.71 & 65.84 &   67.22 & 25.42 & 12.08 & 34.90      \\
\rowcolor{bval} \cellcolor{white} & \bval & 66.15 & 30.11 & 8.715 & 34.99 &   93.16 & 75.49 & 30.18 & 66.27 &   93.57 & 74.92 & 29.20 & 65.89 &   94.87 & 73.38 & 23.76 & 64.00 &   92.85 & 72.69 & 30.53 & 65.35      \\
\rowcolor{bvalvalth} \cellcolor{white}\multirow{-4}{*}{\parbox{1cm}{\textbf{HaS}~\cite{singh2017hide} {\small \emph{(iccv'17)} ResNet50}}} & \bvalvalth & -- & -- & -- & -- &   92.33 & 74.68 & 29.15 & 65.38 &   93.49 & 74.02 & 28.99 & 65.5 &   94.51 & 73.24 & 23.45 & 63.73 &   61.28 & 22.43 & 27.09 & 36.93      \\
\hline

& \btst & 89.48 & 46.59 & 7.490 & 47.85 &   96.35 & 74.64 & 20.81 & 63.93 &   96.35 & 74.64 & 20.81 & 63.93 &   97.20 & 75.37 & 20.29 & 64.28 &   96.85 & 74.14 & 19.41 & 63.46      \\
\rowcolor{btestvalth} \cellcolor{white} & \btestvalth & -- & -- & -- & -- &   95.73 & 73.50 & 20.65 & 63.29 &   96.25 & 73.83 & 20.59 & 63.55 &   96.87 & 74.83 & 19.60 & 63.76 &   88.29 & 33.43 & 9.164 & 43.62      \\
\rowcolor{bval} \cellcolor{white} & \bval & 80.47 & 34.89 & 6.161 & 40.50 &   96.35 & 74.64 & 20.81 & 63.93 &   96.35 & 74.64 & 20.81 & 63.93 &   97.42 & 72.29 & 16.87 & 62.19 &   93.71 & 56.04 & 8.560 & 52.77      \\
\rowcolor{bvalvalth} \cellcolor{white}\multirow{-4}{*}{\parbox{1cm}{\textbf{ACoL}~\cite{zhang2018adversarial} {\small \emph{(cvpr'18)} ResNet50}}} & \bvalvalth & -- & -- & -- & -- &   95.73 & 73.50 & 20.65 & 63.29 &   96.25 & 73.83 & 20.59 & 63.55 &   97.25 & 72.21 & 16.60 & 62.01 &   88.60 & 27.85 & 6.817 & 41.08      \\
\hline

& \btst & 81.08 & 40.97 & 7.973 & 43.34 &   96.47 & 76.63 & 23.14 & 65.41 &   96.47 & 76.63 & 23.14 & 65.41 &   95.35 & 76.06 & 23.16 & 64.85 &   95.46 & 74.99 & 21.86 & 64.10      \\
\rowcolor{btestvalth} \cellcolor{white} & \btestvalth & -- & -- & -- & -- &   96.27 & 76.21 & 22.62 & 65.03 &   96.20 & 76.52 & 23.00 & 65.24 &   95.21 & 75.92 & 22.73 & 64.61 &   76.85 & 27.39 & 10.01 & 38.08      \\
\rowcolor{bval} \cellcolor{white} & \bval & 77.89 & 37.17 & 6.144 & 40.40 &   95.40 & 75.83 & 23.16 & 64.79 &   93.63 & 75.97 & 27.25 & 65.61 &   96.63 & 75.80 & 20.31 & 64.24 &   75.45 & 32.08 & 6.040 & 37.85      \\
\rowcolor{bvalvalth} \cellcolor{white}\multirow{-4}{*}{\parbox{1cm}{\textbf{ADL}~\cite{choe2019attention} {\small \emph{(cvpr'19)} ResNet50}}} & \bvalvalth & -- & -- & -- & -- &   94.87 & 72.73 & 22.69 & 63.43 &   93.04 & 75.85 & 24.02 & 64.30 &   96.08 & 74.73 & 19.57 & 63.46 &   64.03 & 22.21 & 5.505 & 30.58      \\
\hline

& \btst & 90.36 & 52.67 & 12.92 & 51.98 &   91.49 & 65.58 & 19.38 & 58.81 &   91.49 & 65.58 & 19.38 & 58.81 &   91.54 & 65.22 & 18.89 & 58.54 &   76.68 & 45.97 & 17.77 & 46.80      \\
\rowcolor{btestvalth} \cellcolor{white} & \btestvalth & -- & -- & -- & -- &   88.74 & 64.44 & 18.55 & 57.24 &   90.81 & 65.27 & 19.12 & 58.4 &   90.42 & 62.54 & 18.55 & 57.17 &   72.02 & 31.89 & 15.15 & 39.68      \\
\rowcolor{bval} \cellcolor{white} & \bval & 90.36 & 52.67 & 12.92 & 51.98 &   91.49 & 65.58 & 19.38 & 58.81 &   91.49 & 65.58 & 19.38 & 58.81 &   91.54 & 65.22 & 18.89 & 58.54 &   76.68 & 45.97 & 17.77 & 46.80      \\
\rowcolor{bvalvalth} \cellcolor{white}\multirow{-4}{*}{\parbox{1.44cm}{\textbf{CCAM}~\cite{yang2020combinational} {\small \emph{(wacv'20)} VGG}}} & \bvalvalth & -- & -- & -- & -- &   90.12 & 62.44 & 18.77 & 57.11 &   91.37 & 64.84 & 19.27 & 58.49 &   90.42 & 62.54 & 18.55 & 57.17 &   72.02 & 31.89 & 15.15 & 39.68      \\
\hline

& \btst & 85.07 & 51.77 & 20.34 & 52.39 &   99.22 & 90.19 & 55.31 & 81.57 &   99.15 & 90.35 & 55.16 & 81.55 &   98.96 & 89.52 & 55.91 & 81.46 &   98.92 & 88.71 & 51.25 & 79.62      \\
\rowcolor{btestvalth} \cellcolor{white} & \btestvalth & -- & -- & -- & -- &   99.17 & 89.33 & 51.89 & 80.13 &   99.03 & 90.16 & 54.21 & 81.13 &   98.60 & 89.52 & 55.10 & 81.07 &   66.13 & 29.49 & 13.87 & 36.49      \\
\rowcolor{bval} \cellcolor{white} & \bval & 72.41 & 39.17 & 13.72 & 41.76 &   98.96 & 89.52 & 55.91 & 81.46 &   98.96 & 89.52 & 55.91 & 81.46 &   99.11 & 89.16 & 53.50 & 80.58 &   95.27 & 77.39 & 40.90 & 71.18      \\
\rowcolor{bvalvalth} \cellcolor{white}\multirow{-4}{*}{\parbox{1.77cm}{{\footnotesize\textbf{TS-CAM}}~\cite{gao2021ts} {\small \emph{(iccv'21)} DeiT-S}}} & \bvalvalth & -- & -- & -- & -- &   98.56 & 88.85 & 53.40 & 80.27 &   98.87 & 88.90 & 51.58 & 79.78 &   98.44 & 88.95 & 50.98 & 79.45 &   59.52 & 24.19 & 22.19 & 35.30      \\
\hline

& \btst & 62.49 & 30.82 & 8.767 & 34.02 &   99.25 & 91.56 & 56.10 & 82.30 &   99.30 & 92.25 & 58.40 & 83.31 &   99.36 & 92.26 & 58.49 & 83.37 &   99.24 & 91.76 & 56.50 & 82.5      \\
\rowcolor{btestvalth} \cellcolor{white} & \btestvalth & -- & -- & -- & -- &   98.96 & 90.26 & 55.35 & 81.52 &   99.20 & 92.00 & 55.41 & 82.20 &   98.94 & 92.16 & 58.24 & 83.11 &   63.54 & 26.44 & 16.86 & 35.61      \\
\rowcolor{bval} \cellcolor{white} & \bval & 62.49 & 30.82 & 8.767 & 34.02 &   99.25 & 91.56 & 56.10 & 82.30 &   99.30 & 92.25 & 58.40 & 83.31 &   99.36 & 92.26 & 58.49 & 83.37 &   97.39 & 80.16 & 35.86 & 71.13      \\
\rowcolor{bvalvalth} \cellcolor{white}\multirow{-4}{*}{\parbox{1.1cm}{\textbf{SCM}~\cite{bai2022weakly} {\small \emph{(eccv'22)} DeiT-S}}} & \bvalvalth & -- & -- & -- & -- &   98.96 & 90.26 & 55.35 & 81.52 &   99.20 & 92.00 & 55.41 & 82.20 &   98.94 & 92.16 & 58.24 & 83.11 &   63.53 & 22.02 & 10.06 & 31.87      \\
\hline

& \btst & 91.76 & 69.60 & 37.78 & 66.38 &   99.37 & 92.14 & 66.63 & 86.04 &   99.24 & 92.45 & 66.79 & 86.16 &   99.15 & 91.45 & 64.29 & 84.96 &   99.30 & 92.23 & 66.67 & 86.06      \\
\rowcolor{btestvalth} \cellcolor{white} & \btestvalth & -- & -- & -- & -- &   99.36 & 91.66 & 67.34 & 86.12 &   99.03 & 92.26 & 62.82 & 84.70 &   98.49 & 91.00 & 64.03 & 84.50 &   70.65 & 41.62 & 28.04 & 46.77      \\
\rowcolor{bval} \cellcolor{white} & \bval & 92.33 & 70.24 & 39.02 & 67.19 &   99.17 & 91.33 & 65.06 & 85.18 &   99.17 & 91.75 & 68.31 & 86.41 &   99.13 & 89.97 & 60.57 & 83.22 &   99.17 & 91.33 & 65.06 & 85.18      \\
\rowcolor{bvalvalth} \cellcolor{white}\multirow{-4}{*}{\parbox{1.1cm}{\textbf{SAT}~\cite{wuspatial23} {\small \emph{(iccv'23)} DeiT-S}}} & \bvalvalth & -- & -- & -- & -- &   98.87 & 90.86 & 65.06 & 84.93 &   98.79 & 91.71 & 66.44 & 85.64 &   97.82 & 89.93 & 60.40 & 82.71 &   77.59 & 28.04 & 14.68 & 40.10      \\
\hline

\end{tabular}
}
\captionof{table}{
Test-set \newmaxboxacc that is average of three threshold \iouthirty, \ioufifty, \iouseventy (here \newmaxboxacc is aerage at three threhold) along with the it of WSOL models with different selection criteria on \cubs. The \textit{select} column presents (i) BT and BV indicate model selection based on hyperparameter configurations using the test set and validation set, respectively; (ii) TT and VT indicate that the threshold \(\tau\) is selected using either the test set or validation set. For model selection on the validation set, we consider the GT as a reference, a selection based on the \classification performance and the three different pseudo-bboxes generation proposed in this work: RPN, CLIP and SS. Our results for models selected with pseudo-bboxes are comparable to those of GT.
}
\label{wacveval:tab:supp-results_cub_detialed}
\end{table*}

\begin{table*}[!t]
\centering
\resizebox{1\linewidth}{!}{%
\begin{tabular}{|l||c||c|c|c|c||c|c|c|c||c|c|c|c||c|c|c|c||c|c|c|c|c|}
\hline
						& & \multicolumn{4}{c|}{\textbf{CL}}  & \multicolumn{4}{c|}{\textbf{GT}}  & \multicolumn{4}{c|}{\textbf{RPN}}  & \multicolumn{4}{c|}{\textbf{CLIP}}  & \multicolumn{4}{c|}{\textbf{SS}} \\\hline

\textbf{Method} & \textbf{Select} & \iouthirty  & \ioufifty  & \iouseventy  &  {\footnotesize\newmaxboxaccs}  & \iouthirty  & \ioufifty  & \iouseventy  &  {\footnotesize\newmaxboxaccs}  & \iouthirty  & \ioufifty  & \iouseventy  &  {\footnotesize\newmaxboxaccs}  & \iouthirty  & \ioufifty  & \iouseventy  &  {\footnotesize\newmaxboxaccs}  & \iouthirty  & \ioufifty  & \iouseventy  &  {\footnotesize\newmaxboxaccs} \\\hline

& \btst & 73.16 & 46.29 & 20.86 & 46.77 &   81.99 & 64.06 & 40.22 & 62.09 &   81.59 & 63.60 & 39.55 & 61.58 &   82.02 & 63.90 & 39.98 & 61.96 &   81.59 & 63.60 & 39.55 & 61.58      \\
\rowcolor{btestvalth} \cellcolor{white} & \btestvalth & -- & -- & -- & -- &   81.98 & 63.90 & 39.59 & 61.82 &   79.49 & 62.89 & 37.64 & 60.00 &   81.09 & 63.88 & 39.85 & 61.60 &   68.59 & 46.97 & 39.84 & 51.80      \\
\rowcolor{bval} \cellcolor{white} & \bval & 73.81 & 46.83 & 21.26 & 47.30 &   81.99 & 64.06 & 40.22 & 62.09 &   81.59 & 63.60 & 39.55 & 61.58 &   82.06 & 64.01 & 40.02 & 62.03 &   81.59 & 63.60 & 39.55 & 61.58      \\
\rowcolor{bvalvalth} \cellcolor{white}\multirow{-4}{*}{\parbox{1cm}{\textbf{CAM}~\cite{zhou2016learning} {\small \emph{(cvpr'16)} ResNet50}}} & \bvalvalth & -- & -- & -- & -- &   81.88 & 63.88 & 39.94 & 61.9 &   79.49 & 62.89 & 37.64 & 60.00 &   81.09 & 63.88 & 39.85 & 61.60 &   70.13 & 45.91 & 39.51 & 51.84      \\
\hline

& \btst & 65.09 & 37.82 & 16.15 & 39.68 &   81.3 & 63.77 & 42.08 & 62.38 &   81.54 & 63.94 & 41.7 & 62.39 &   81.3 & 63.77 & 42.08 & 62.38 &   81.43 & 63.30 & 39.26 & 61.33      \\
\rowcolor{btestvalth} \cellcolor{white} & \btestvalth & -- & -- & -- & -- &   81.16 & 63.86 & 41.67 & 62.23 &   78.59 & 62.08 & 37.35 & 59.34 &   80.35 & 63.28 & 40.95 & 61.52 &   68.92 & 49.02 & 37.13 & 51.69      \\
\rowcolor{bval} \cellcolor{white} & \bval & 72.59 & 45.80 & 21.01 & 46.46 &   81.54 & 63.94 & 41.7 & 62.39 &   81.55 & 63.34 & 39.77 & 61.55 &   81.82 & 63.54 & 39.56 & 61.63 &   81.43 & 63.30 & 39.26 & 61.33      \\
\rowcolor{bvalvalth} \cellcolor{white}\multirow{-4}{*}{\parbox{1cm}{\textbf{HaS}~\cite{singh2017hide} {\small \emph{(iccv'17)} ResNet50}}} & \bvalvalth & -- & -- & -- & -- &   81.16 & 63.86 & 41.67 & 62.23 &   78.59 & 62.08 & 37.35 & 59.34 &   80.36 & 63.06 & 39.21 & 60.87 &   68.56 & 48.21 & 39.23 & 52.0      \\
\hline

& \btst & 77.99 & 50.46 & 22.76 & 50.40 &   81.34 & 62.93 & 38.23 & 60.83 &   81.89 & 62.75 & 37.56 & 60.73 &   82.21 & 63.45 & 39.13 & 61.59 &   82.14 & 62.92 & 37.50 & 60.85      \\
\rowcolor{btestvalth} \cellcolor{white} & \btestvalth & -- & -- & -- & -- &   81.83 & 63.48 & 39.11 & 61.47 &   79.94 & 61.61 & 36.75 & 59.43 &   80.38 & 63.47 & 39.11 & 60.98 &   76.95 & 52.38 & 33.11 & 54.14      \\
\rowcolor{bval} \cellcolor{white} & \bval & 77.99 & 50.46 & 22.76 & 50.40 &   82.23 & 63.70 & 39.03 & 61.65 &   81.89 & 62.75 & 37.56 & 60.73 &   82.14 & 62.93 & 37.5 & 60.85 &   82.14 & 62.92 & 37.50 & 60.85      \\
\rowcolor{bvalvalth} \cellcolor{white}\multirow{-4}{*}{\parbox{1cm}{\textbf{ACoL}~\cite{zhang2018adversarial} {\small \emph{(cvpr'18)} ResNet50}}} & \bvalvalth & -- & -- & -- & -- &   81.83 & 63.48 & 39.11 & 61.47 &   80.45 & 61.31 & 35.89 & 59.21 &   81.37 & 62.80 & 37.43 & 60.53 &   76.95 & 52.38 & 33.11 & 54.14      \\
\hline

& \btst & 67.71 & 39.80 & 16.84 & 41.44 &   82.67 & 65.11 & 41.63 & 63.13 &   82.51 & 65.97 & 41.70 & 63.39 &   82.51 & 65.11 & 41.63 & 63.08 &   82.64 & 65.19 & 41.92 & 63.25      \\
\rowcolor{btestvalth} \cellcolor{white} & \btestvalth & -- & -- & -- & -- &   82.66 & 65.2 & 41.49 & 63.11 &   80.68 & 64.29 & 37.49 & 60.82 &   81.65 & 65.04 & 41.33 & 62.67 &   69.47 & 49.61 & 41.59 & 53.55      \\
\rowcolor{bval} \cellcolor{white} & \bval & 70.90 & 44.23 & 20.57 & 45.23 &   82.67 & 65.28 & 41.88 & 63.27 &   82.64 & 65.18 & 41.92 & 63.24 &   82.69 & 65.32 & 41.79 & 63.26 &   82.58 & 65.06 & 41.28 & 62.97      \\
\rowcolor{bvalvalth} \cellcolor{white}\multirow{-4}{*}{\parbox{1cm}{\textbf{ADL}~\cite{choe2019attention} {\small \emph{(cvpr'19)} ResNet50}}} & \bvalvalth & -- & -- & -- & -- &   82.66 & 65.2 & 41.49 & 63.11 &   80.41 & 62.30 & 37.63 & 60.11 &   81.65 & 65.04 & 41.33 & 62.67 &   72.96 & 47.09 & 41.32 & 53.79      \\
\hline

& \btst & 73.84 & 49.80 & 24.99 & 49.54 &   77.84 & 60.63 & 38.39 & 58.95 &   77.84 & 60.63 & 38.4 & 58.95 &   77.83 & 60.63 & 38.39 & 58.95 &   72.34 & 52.72 & 31.22 & 52.09      \\
\rowcolor{btestvalth} \cellcolor{white} & \btestvalth & -- & -- & -- & -- &   77.72 & 60.52 & 38.02 & 58.75 &   77.19 & 60.52 & 37.76 & 58.49 &   77.44 & 60.56 & 38.17 & 58.72 &   67.94 & 46.63 & 31.16 & 48.57      \\
\rowcolor{bval} \cellcolor{white} & \bval & 73.84 & 49.80 & 24.99 & 49.54 &   77.84 & 60.63 & 38.39 & 58.95 &   77.84 & 60.63 & 38.4 & 58.95 &   77.83 & 60.63 & 38.39 & 58.95 &   72.34 & 52.72 & 31.22 & 52.09      \\
\rowcolor{bvalvalth} \cellcolor{white}\multirow{-4}{*}{\parbox{1.44cm}{\textbf{CCAM}~\cite{yang2020combinational} {\small \emph{(wacv'20)} VGG}}} & \bvalvalth & -- & -- & -- & -- &   77.72 & 60.52 & 38.02 & 58.75 &   77.19 & 60.52 & 37.76 & 58.49 &   77.44 & 60.56 & 38.17 & 58.72 &   67.94 & 46.63 & 31.16 & 48.57      \\
\hline

& \btst & 12.45 & 6.476 & 2.144 & 7.023 &   82.91 & 66.75 & 43.77 & 64.47 &   82.91 & 66.75 & 43.77 & 64.47 &   82.91 & 66.75 & 43.77 & 64.47 &   82.56 & 66.17 & 42.87 & 63.86      \\
\rowcolor{btestvalth} \cellcolor{white} & \btestvalth & -- & -- & -- & -- &   82.74 & 66.65 & 43.77 & 64.38 &   79.95 & 65.12 & 41.18 & 62.08 &   80.42 & 65.32 & 43.55 & 63.09 &   67.53 & 47.63 & 35.03 & 50.06      \\
\rowcolor{bval} \cellcolor{white} & \bval & 12.45 & 6.476 & 2.144 & 7.023 &   82.91 & 66.75 & 43.77 & 64.47 &   82.52 & 66.14 & 42.94 & 63.86 &   82.91 & 66.75 & 43.77 & 64.47 &   73.60 & 55.7 & 35.03 & 54.77      \\
\rowcolor{bvalvalth} \cellcolor{white}\multirow{-4}{*}{\parbox{1.77cm}{{\footnotesize\textbf{TS-CAM}}~\cite{gao2021ts} {\small \emph{(iccv'21)} DeiT-S}}} & \bvalvalth & -- & -- & -- & -- &   82.74 & 66.65 & 43.77 & 64.38 &   80.68 & 64.97 & 42.80 & 62.81 &   80.42 & 65.32 & 43.55 & 63.09 &   66.87 & 46.8 & 34.08 & 49.25      \\
\hline

& \btst & 74.53 & 47.55 & 17.72 & 46.6 &   80.60 & 61.76 & 34.44 & 58.93 &   80.60 & 61.75 & 34.44 & 58.93 &   80.60 & 61.75 & 34.44 & 58.93 &   79.42 & 59.76 & 32.95 & 57.37      \\
\rowcolor{btestvalth} \cellcolor{white} & \btestvalth & -- & -- & -- & -- &   80.48 & 61.76 & 34.42 & 58.88 &   78.61 & 61.15 & 30.44 & 56.73 &   79.90 & 61.38 & 34.27 & 58.51 &   68.54 & 49.27 & 30.88 & 49.56      \\
\rowcolor{bval} \cellcolor{white} & \bval & 74.53 & 47.55 & 17.72 & 46.6 &   80.60 & 61.76 & 34.44 & 58.93 &   80.60 & 61.75 & 34.44 & 58.93 &   80.60 & 61.75 & 34.44 & 58.93 &   79.42 & 59.76 & 32.95 & 57.37      \\
\rowcolor{bvalvalth} \cellcolor{white}\multirow{-4}{*}{\parbox{1.1cm}{\textbf{SCM}~\cite{bai2022weakly} {\small \emph{(eccv'22)} DeiT-S}}} & \bvalvalth & -- & -- & -- & -- &   80.48 & 61.76 & 34.42 & 58.88 &   79.50 & 60.21 & 34.39 & 58.03 &   79.90 & 61.38 & 34.27 & 58.51 &   68.54 & 49.27 & 30.88 & 49.56      \\
\hline

& \btst & 79.54 & 64.59 & 37.47 & 60.53 &   82.92 & 70.12 & 52.93 & 68.65 &   82.33 & 67.08 & 47.01 & 65.47 &   82.92 & 70.13 & 52.94 & 68.66 &   82.92 & 70.13 & 52.94 & 68.66      \\
\rowcolor{btestvalth} \cellcolor{white} & \btestvalth & -- & -- & -- & -- &   82.91 & 69.46 & 52.82 & 68.39 &   79.96 & 64.78 & 45.18 & 63.30 &   81.19 & 69.05 & 52.25 & 67.49 &   73.13 & 57.06 & 51.02 & 60.40      \\
\rowcolor{bval} \cellcolor{white} & \bval & 80.09 & 66.17 & 43.52 & 63.26 &   82.92 & 70.12 & 52.93 & 68.65 &   82.33 & 67.08 & 47.01 & 65.47 &   82.92 & 70.13 & 52.94 & 68.66 &   82.92 & 70.13 & 52.94 & 68.66      \\
\rowcolor{bvalvalth} \cellcolor{white}\multirow{-4}{*}{\parbox{1.1cm}{\textbf{SAT}~\cite{wuspatial23} {\small \emph{(iccv'23)} DeiT-S}}} & \bvalvalth & -- & -- & -- & -- &   82.91 & 69.46 & 52.82 & 68.39 &   79.96 & 64.78 & 45.18 & 63.30 &   81.19 & 69.05 & 52.25 & 67.49 &   73.13 & 57.06 & 51.02 & 60.40      \\
\hline

\end{tabular}
}
\captionof{table}{
Test-set \newmaxboxacc that is average of three threshold \iouthirty, \ioufifty, \iouseventy (here \newmaxboxacc is aerage at three threhold) along with the it of WSOL models with different selection criteria on \ilsvrc. The \textit{select} column presents (i) BT and BV indicate model selection based on hyperparameter configurations using the test set and validation set, respectively; (ii) TT and VT indicate that the threshold \(\tau\) is selected using either the test set or validation set. For model selection on the validation set, we consider the GT as a reference, a selection based on the \classification performance and the three different pseudo-bboxes generation proposed in this work: RPN, CLIP and SS. Our results for models selected with pseudo-bboxes are comparable to those of GT.
\vspace{-0.5cm}
}
\label{wacveval:tab:supp-results_ilsvrc_detialed}
\end{table*}

\begin{figure*}[!t]
\begin{center}
\includegraphics[width=1\linewidth,trim=0 0 0 0, clip]{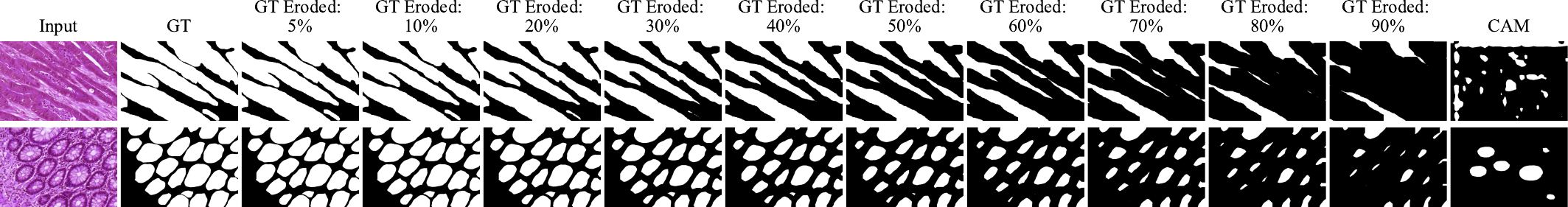}
\end{center}
\vspace{-0.44cm}
\caption{Illustration GT masks, noisy masks at various erosion levels.
\vspace{-0.21cm}
}\label{wacveval:fig:noisy_medical_ann}
\end{figure*}

\section{Experiments with medical datasets}
\label{wacveval:sec:medicalexp}

To show the robustness and generalizability of our evaluation protocol across datasets with varying characteristics, we extended our protocol to the task of localization in histology images—a particularly challenging problem due to its complexity and sensitivity, especially for non-experts attempting to identify regions of interest. For this, we employed two histology image datasets and first simulated noisy masks by applying erosion to the GT masks. This approach allowed us to assess the protocol's robustness against different levels of noise. Additionally for realistic setup, we extracted pseudo-masks from the activation maps of a pre-trained classifier, thereby evaluating the protocol's performance in a realistic setting. Our results indicate that the evaluation protocol consistently maintains performance across varying noise levels, confirming its robustness in both simulated and real-world scenarios.

\noindent\textbf{Datasets.}
We employed two additional datasets to show the robustness of our proposed protocol; \textit{(i) \glas dataset} is collected for the diagnosis of colon cancer and comprises 165 images derived from 16 Hematoxylin and Eosin (H\&E) stained slides. It includes pixel-level and image-level annotations (benign or malignant). The dataset is divided into 67 images for training, 18 for validation, and 80 for testing. Our evaluation follows the protocol established by \cite{rony2023deep}, where three fully supervised examples per class are used for best model over \localization. \textit{(ii) \camelyon dataset} is a patch-based benchmark extracted using Camelyon16 dataset, which consists of 399 whole slide images with two classes (normal and metastatic) used for detecting metastases in H\&E-stained tissue sections of sentinel lymph nodes from breast cancer patients. Following the extraction protocol outlined in \cite{rony2023deep}, patches of size 512$\times$512 are annotated at both image and pixel levels. This dataset comprises a total of 48,870 images, with 24,348 for training, 8,850 for validation, and 15,664 for testing. From the validation set, six fully supervised examples per class are randomly selected to determine best model, as suggested by \cite{rony2023deep}.

\noindent\textbf{Evaluation measures.} The \glas and \camelyon datasets provide pixel-wise annotations (masks) rather than bboxes, necessitating the use of pixel average precision (\pxap) \cite{choe2020evaluating, rony2023deep} to evaluate localization accuracy. Following the standard WSOL pipeline, we first employ min-max normalization on these activation maps and apply various thresholds to the activation maps for producing localization maps.

\noindent\textbf{Generation of noisy and pseudo-GT masks.} 
The objective of this study is to evaluate the robustness of the proposed model selection techniques under conditions of noisy masks. To this end, we perturb the GT masks through erosion, simulating scenarios with noisy or imprecise GT annotations commonly encountered in practical applications. The noisy GT masks are generated by applying erosion with varying filter sizes and iterations until the masks are degraded to a specified extent, corresponding to eleven predefined noise levels, as detailed in Tab.\ref{wacveval:tab:noisy_medical_ann} and examples of noisy masks are illustrated in Fig.\ref{wacveval:fig:noisy_medical_ann}.

\noindent\textbf{Results.} Tab.\ref{wacveval:tab:noisy_medical_ann} compares the performance of models (ResNet50) trained with pseudo-masks and generated pseudo-masks against those trained with GT masks, demonstrating the robustness of our evaluation protocol across varying noise levels. The results include localization performance for models using pseudo-masks generated by a pre-trained InceptionV3 classifier, an architecture distinct from the one used for training our models. Despite significant increases in mask error across different noise levels on both the \glas and \camelyon datasets, localization performance remains relatively stable, even at the highest noise level. Additionally, pseudo-masks generated from a different architecture show reasonable localization accuracy, further validating the generalizability of our evaluation protocol to diverse types of annotation noise, and underscoring its robustness in practical settings.

\setlength{\tabcolsep}{6pt}
\definecolor{light-gray}{gray}{0.9}
\begin{table}[!h]
\centering
\resizebox{0.9\columnwidth}{!}{%
\begin{tabular}{|c|cc|cc|}
\cline{2-5}
\multicolumn{1}{c|}{}   & \multicolumn{2}{c|}{\large\textbf{\glas}} & \multicolumn{2}{c|}{\large	\textbf{\camelyon}} \\ \hline
\textbf{Pseudo Mask} &
  \multicolumn{1}{c|}{\textbf{\begin{tabular}[c]{@{}c@{}}Mask Error\\ 1-AUC\end{tabular}}} &
  \multicolumn{1}{c|}{\textbf{\begin{tabular}[c]{@{}c@{}}LOC\\ \pxap\end{tabular}}} &
  \multicolumn{1}{c|}{\textbf{\begin{tabular}[c]{@{}c@{}}Mask Error\\ 1-AUC\end{tabular}}} &
  \multicolumn{1}{c|}{\textbf{\begin{tabular}[c]{@{}c@{}}LOC\\ \pxap\end{tabular}}} \\ \hline\hline
\rowcolor{light-gray}  GT-Mask           & \multicolumn{1}{c|}{0.0}   & 69.78 & \multicolumn{1}{c|}{0.0}       & 30.55    \\ \hline
\multicolumn{1}{|c|}{\textbf{\begin{tabular}[c]{@{}c@{}}Perturbed GT-Mask \\with erosion level:\end{tabular}}}           & \multicolumn{4}{c|}{}  \\ \hline
5\%           & \multicolumn{1}{c|}{1.86}  & 70.09 & \multicolumn{1}{c|}{1.64}      & 29.74    \\ \hline
10\%          & \multicolumn{1}{c|}{4.30}   & 70.3  & \multicolumn{1}{c|}{4.23}      & 38.87    \\ \hline
20\%         & \multicolumn{1}{c|}{9.22}  & 70.40  & \multicolumn{1}{c|}{9.29}      & 29.74    \\ \hline
30\%        & \multicolumn{1}{c|}{14.61} & 70.02 & \multicolumn{1}{c|}{14.26}     & 30.55    \\ \hline
40\%        & \multicolumn{1}{c|}{19.15} & 69.83 & \multicolumn{1}{c|}{19.17}     & 29.74    \\ \hline
50\%        & \multicolumn{1}{c|}{24.34} & 70.58 & \multicolumn{1}{c|}{24.22}     & 29.74    \\ \hline
60\%        & \multicolumn{1}{c|}{29.18} & 67.68 & \multicolumn{1}{c|}{29.14}     & 29.74    \\ \hline
70\%        & \multicolumn{1}{c|}{34.16} & 68.38 & \multicolumn{1}{c|}{34.11}     & 29.74    \\ \hline
80\%        & \multicolumn{1}{c|}{39.21} & 67.98 & \multicolumn{1}{c|}{39.17}     & 29.74    \\ \hline
90\%        & \multicolumn{1}{c|}{44.12} & 68.65 & \multicolumn{1}{c|}{44.14}     & 29.74    \\ \hline
\rowcolor{light-gray}  CAM \cite{zhou2016learning} Pseudo-Mask & \multicolumn{1}{c|}{46.54} & 66.41 & \multicolumn{1}{c|}{28.10}      & 29.74    \\ \hline
\end{tabular}%
}
\caption{Impact of varying levels of erosion in GT masks on \localization performance (\pxap) compared to GT-mask and pseudo-masks. It shows that despite substantial increases in mask error (1-AUC) across different noise levels, the \localization performance remains stable, particularly when using pseudo-masks, highlighting the robustness and generalizability of our evaluation protocol across different domains.}
\label{wacveval:tab:noisy_medical_ann}
\end{table}

%% file: aux_files/main.bib
@String(IJCV  = {Int. J. Comput. Vis.})

@String(CVPR  = {IEEE Conf. Comput. Vis. Pattern Recog.})

@String(ICCV  = {Int. Conf. Comput. Vis.})

@String(ECCV  = {Eur. Conf. Comput. Vis.})

@String(WACV  = {Win. Conf. App. Comp. Vis.})

@String(NeurIPS = {Adv. Neural Inform. Process. Syst.})

@String(ICML  = {Int. Conf. Mach. Learn.})

@String(ICLR  = {Int. Conf. Learn. Represent.})

@String(BMVC  = {Brit. Mach. Vis. Conf.})

@String(IJCV  = {IJCV})

@String(CVPR  = {CVPR})

@String(ICCV  = {ICCV})

@String(ECCV  = {ECCV})

@String(WACV  = {WACV})

@String(NeurIPS = {NeurIPS})

@String(ICML  = {ICML})

@String(ICLR  = {ICLR})

@String(BMVC  =	{BMVC})

@InProceedings{belharbi2022fcam,
  title        = {{F-CAM}: Full Resolution Class Activation Maps via Guided Parametric Upscaling},
  author       = {S. Belharbi and 
                  A. Sarraf and 
                  M. Pedersoli and 
                  I. Ben Ayed and 
                  L. McCaffrey and 
                  E. Granger},
  booktitle = {WACV},
  year         = {2022}
}

@inproceedings{choe2019attention,
  title        = {Attention-based dropout layer for weakly supervised object localization},
  author       = {J. Choe and 
                  H. Shim},
  booktitle    = {CVPR},
  year         = {2019}
}

@inproceedings{singh2017hide,
  title        = {Hide-and-seek: Forcing a network to be meticulous for weakly-supervised object and action localization},
  author       = {K. Kumar Singh and Y. Jae Lee},
  booktitle    = {ICCV},
  year         = {2017},
}

@inproceedings{yang2020combinational,
  title        = {Combinational class activation maps for weakly supervised object localization},
  author       = {S. Yang and 
                  Y. Kim and 
                  Y. Kim and 
                  C. Kim},
  booktitle    = {WACV},
  year         = {2020}
}

@article{zhang2018top,
  author       ={J. Zhang and 
                S. Adel Bargal and 
                Z. Lin and 
                J. Brandt and 
                X. Shen and 
                S. Sclaroff},
  title        = {Top-down neural attention by excitation backprop},
  journal      = {IJCV},
  volume       = {126},
  number       = {10},
  pages        = {1084--1102},
  year         = {2018}
}

@inproceedings{zhou2016learning,
  title        = {Learning deep features for discriminative localization},
  author       = {B. Zhou and 
                  A. Khosla and 
                  A. Lapedriza and 
                  A. Oliva and 
                  A. Torralba},
  booktitle    = {CVPR},
  year         = {2016}
}

@inproceedings{zhang2018adversarial,
  title        = {Adversarial complementary learning for weakly supervised object localization},
  author       = {X. Zhang and 
                  Y. Wei and 
                  J. Feng and 
                  Y. Yang and 
                  T. Huang},
  booktitle    = {CVPR},
  year         = {2018}
}

@inproceedings{gao2021ts,
  title        = {Ts-cam: Token semantic coupled attention map for weakly supervised object localization},
  author       = {W. Gao and 
                  F. Wan and 
                  X. Pan and 
                  Z. Peng and 
                  Q. Tian and 
                  Z. Han and 
                  B. Zhou and 
                  Q. Ye},
  booktitle    = {ICCV},
  year         = {2021}
}

@article{bai2022weakly,
  title        = {Weakly Supervised Object Localization via Transformer with Implicit Spatial Calibration},
  author       = {H. Bai and 
                  R. Zhang and 
                  J. Wang and 
                  X. Wan},
  journal      = {ECCV},
  year         = {2022}
}

@inproceedings{wuspatial23,
  author       = {P. Wu and
                  W. Zhai and
                  Y. Cao and
                  J. Luo and
                  ZJ. Zha},
  title        = {Spatial-Aware Token for Weakly Supervised Object Localization},
  booktitle    = {ICCV},
  year         = {2023}
}

@inproceedings{choe2020evaluating,
  title        = {Evaluating weakly supervised object localization methods right},
  author       = {J. Choe and 
                  S. Oh and 
                  S. Lee and 
                  S. Chun and 
                  Z. Akata and H. Shim},
  booktitle    = {CVPR},
  year         = {2020}
}

@techreport{wah2011caltech,
  title        = {The Caltech-UCSD Birds-200-2011 Dataset},
  author       = {C. Wah and 
                  S. Branson and
                  W. Steve and
                  P. Peter and 
                  S. Belongie},
  year         = {2011},
  institution  = {California Institute of Technology},
  number       = {CNS-TR-2011-001}
}

@inproceedings{imagenet_cvpr09,
  author       = {J. Deng and 
                  W. Dong and 
                  R. Socher and 
                  L.J. Li and 
                  K. Li and 
                  L. Fei-Fei},
  title        = {ImageNet: A Large-Scale Hierarchical Image Database},
  booktitle    = CVPR,
  year         = {2009},
}

@inproceedings{recht2019imagenet,
  title        = {Do imagenet classifiers generalize to imagenet?},
  author       = {Recht, Benjamin and Roelofs, Rebecca and Schmidt, Ludwig and Shankar, Vaishaal},
  booktitle    = {ICML},
  year         = {2019},
}

@article{uijlings13,
  title={Selective search for object recognition},
  author={Uijlings, J.RR and Van De Sande, K.EA and Gevers, T. and Smeulders, A.WM},
  journal={IJCV},
  volume={104},
  pages={154--171},
  year={2013}
}

@article{ren15,
  title={Faster r-cnn: Towards real-time object detection with region proposal networks},
  author={Ren, S. and He, K. and Girshick, R. and Sun, J.},
  journal={NeurIPS},
  year={2015}
}

@inproceedings{lin23,
  title={{CLIP} is also an efficient segmenter: A text-driven approach for weakly supervised semantic segmentation},
  author={Lin, Y. and Chen, M. and Wang, W. and Wu, B. and Li, K. and Lin, B. and Liu, H. and He, X},
  booktitle={CVPR},
  year={2023}
}

@inproceedings{Hosang14,
  author       = {J.H. Hosang and
                  R. Benenson and
                  B. Schiele},
  title        = {How good are detection proposals, really?},
  booktitle    = {BMVC},
  year         = {2014}
}

@inproceedings{Oliver18,
  author       = {A. Oliver and
                  A. Odena and
                  C. Raffel and
                  E. Dogus Cubuk and
                  I.J. Goodfellow},
  title        = {Realistic Evaluation of Deep Semi-Supervised Learning Algorithms},
  booktitle    = {NeurIPS},
  year         = {2018}
}

@inproceedings{Su21,
  author       = {J.{-}C. Su and
                  Z. Cheng and
                  S. Maji},
  title        = {A Realistic Evaluation of Semi-Supervised Learning for Fine-Grained
                  Classification},
  booktitle    = {CVPR},
  year         = {2021}
}

@inproceedings{salvador22,
title={A Reproducible and Realistic Evaluation of Partial Domain Adaptation Methods},
author={T. Salvador and K. FATRAS and I. Mitliagkas and A. M Oberman},
booktitle={NeurIPS Workshop on Distribution Shifts: Connecting Methods and Applications},
year={2022}
}

@inproceedings{Dinu23,
  author       = {M.{-}C. Dinu and
                  M. Holzleitner and
                  M. Beck and
                  H. D. Nguyen and
                  A. Huber and
                  H. Eghbal{-}zadeh and
                  B. A. Moser and
                  S. V. Pereverzyev and
                  S. Hochreiter and
                  W. Zellinger},
  title        = {Addressing Parameter Choice Issues in Unsupervised Domain Adaptation
                  by Aggregation},
  booktitle    = {ICLR},
  year         = {2023}
}

@inproceedings{Ericsson23,
  author       = {L. Ericsson and
                  D. Li and
                  T. M. Hospedales},
  title        = {Better Practices for Domain Adaptation},
  booktitle    = {AutoML},
  year         = {2023}
}

@article{Yang23,
  author       = {J. Yang and
                  H. Qian and
                  Y. Xu and
                  L. Xie},
  title        = {Can We Evaluate Domain Adaptation Models Without Target-Domain Labels?},
  journal      = {CoRR},
  volume       = {abs/2305.18712},
  year         = {2023}
}

@inproceedings{You19,
  author       = {K. You and
                  X. Wang and
                  M. Long and
                  M. I. Jordan},
  editor       = {Kamalika Chaudhuri and
                  Ruslan Salakhutdinov},
  title        = {Towards Accurate Model Selection in Deep Unsupervised Domain Adaptation},
  booktitle    = {ICML},
  year         = {2019}
}

@inproceedings{Saito21,
  author       = {K. Saito and
                  D. Kim and
                  P. Teterwak and
                  S. Sclaroff and
                  T. Darrell and
                  K. Saenko},
  title        = {Tune it the Right Way: Unsupervised Validation of Domain Adaptation
                  via Soft Neighborhood Density},
  booktitle    = {ICCV},
  year         = {2021}
}

@article{morgan89,
  title={Generalization and parameter estimation in feedforward nets: Some experiments},
  author={Morgan, N. and Bourlard, H.},
  journal={NeurIPS},
  volume={2},
  year={1989}
}

@article{Finnoff93,
  author       = {W. Finnoff and
                  F. Hergert and
                  H.{-}G. Zimmermann},
  title        = {Improving model selection by nonconvergent methods},
  journal      = {Neural Networks},
  volume       = {6},
  number       = {6},
  pages        = {771--783},
  year         = {1993}
}

@inproceedings{Lodwich09,
  author       = {A. Lodwich and
                  Y. Rangoni and
                  T. M. Breuel},
  title        = {Evaluation of robustness and performance of Early Stopping Rules with
                  Multi Layer Perceptrons},
  booktitle    = {IJCNN},
  year         = {2009}
}

@article{cheplygina19,
  title={Not-so-supervised: a survey of semi-supervised, multi-instance, and transfer learning in medical image analysis},
  author={Cheplygina, V. and de Bruijne, M. and Pluim, J.PW},
  journal={Medical image analysis},
  volume={54},
  pages={280--296},
  year={2019}
}

@article{zhou18,
  title={A brief introduction to weakly supervised learning},
  author={Zhou, Z.-H.},
  journal={National science review},
  volume={5},
  number={1},
  pages={44--53},
  year={2018}
}

@inproceedings{caron21,
  title={Emerging properties in self-supervised vision transformers},
  author={Caron, M. and Touvron, H. and Misra, I. and J{\'e}gou, H. and Mairal, J. and Bojanowski, P. and Joulin, A.},
  booktitle={ICCV},
  pages={9650--9660},
  year={2021}
}

@article{chen23,
  author       = {W. Chen and
                  M. L. Cummings},
  title        = {Subjectivity in Unsupervised Machine Learning Model Selection},
  journal      = {CoRR},
  volume       = {abs/2309.00201},
  year         = {2023}
}

@article{Borji19,
  author       = {A. Borji},
  title        = {Pros and cons of {GAN} evaluation measures},
  journal      = {Computer Vision and Image Understanding},
  volume       = {179},
  pages        = {41--65},
  year         = {2019}
}

@inproceedings{Zhang18,
  author       = {R. Zhang and
                  P. Isola and
                  A. A. Efros and
                  E. Shechtman and
                  O. Wang},
  title        = {The Unreasonable Effectiveness of Deep Features as a Perceptual Metric},
  booktitle    = {CVPR},
  year         = {2018}
}

@inproceedings{Salimans16,
  author       = {T. Salimans and
                  I.J. Goodfellow and
                  W. Zaremba and
                  V. Cheung and
                  A. Radford and
                  X. Chen},
  title        = {Improved Techniques for Training GANs},
  booktitle    = {NeurIPS},
  year         = {2016}
}

@inproceedings{Heusel17,
  author       = {M. Heusel and
                  H. Ramsauer and
                  T. Unterthiner and
                  B. Nessler and
                  S. Hochreiter},
  title        = {GANs Trained by a Two Time-Scale Update Rule Converge to a Local Nash
                  Equilibrium},
  booktitle    = {NeurIPS},
  year         = {2017}
}

@article{fang23,
  author       = {Y. Fang and
                  P.{-}T. Yap and
                  W. Lin and
                  H. Zhu and
                  M. Liu},
  title        = {Source-Free Unsupervised Domain Adaptation: {A} Survey},
  journal      = {CoRR},
  volume       = {abs/2301.00265},
  year         = {2023}
}

@INPROCEEDINGS{meethal22,
  author={Meethal, A. and Pedersoli, M. and Zhu, Z. and Romero, F.P. and Granger, E.},
  booktitle={IJCNN}, 
  title={Semi-Weakly Supervised Object Detection by Sampling Pseudo Ground-Truth Boxes}, 
  year={2022}
}

@ARTICLE{otsuthresh79,  
author={Otsu, N.},  
journal={IEEE Transactions on Systems, Man, and Cybernetics},   
title={A Threshold Selection Method from Gray-Level Histograms},  
year={1979},  
volume={9},  
number={1},  
pages={62-66}
}

@inproceedings{Shi12,
  author       = {Y. Shi and
                  F. Sha},
  title        = {Information-Theoretical Learning of Discriminative Clusters for Unsupervised
                  Domain Adaptation},
  booktitle    = {ICLM},
  year         = {2012}
}

@inproceedings{vaithyanathan99,
  author={Vaithyanathan, S. and Dom, B.},
  title={Generalized model selection for unsupervised learning in high dimensions},
  booktitle    = {NeurIPS},
  year         = {1999}
}

@article{Mahsereci17,
  author       = {M. Mahsereci and
                  L. Balles and
                  C. Lassner and
                  P. Hennig},
  title        = {Early Stopping without a Validation Set},
  journal      = {CoRR},
  volume       = {abs/1703.09580},
  year         = {2017}
}

@article{li20,
  author       = {W. Li and
                  C. Geng and
                  S. Chen},
  title        = {Leave Zero Out: Towards a No-Cross-Validation Approach for Model Selection},
  journal      = {CoRR},
  volume       = {abs/2012.13309},
  year         = {2020}
}

@inproceedings{Bonet21,
  author       = {D. Bonet and
                  A. Ortega and
                  J. Ruiz Hidalgo and
                  S. Shekkizhar},
  title        = {Channel-Wise Early Stopping without a Validation Set via {NNK} Polytope
                  Interpolation},
  booktitle    = {APSIPA},
  year         = {2021}
}

@article{Shekkizhar20,
  author       = {S. Shekkizhar and
                  A. Ortega},
  title        = {DeepNNK: Explaining deep models and their generalization using polytope
                  interpolation},
  journal      = {CoRR},
  volume       = {abs/2007.10505},
  year         = {2020}
}

@inproceedings{duvenaud16,
  title={Early stopping as nonparametric variational inference},
  author={Duvenaud, D. and Maclaurin, D. and Adams, R.},
  booktitle={Artificial intelligence and statistics},
  pages={1070--1077},
  year={2016}
}

@inproceedings{yuan24,
title={Early Stopping Against Label Noise Without Validation Data},
author={S. Yuan and L. Feng and T. Liu},
booktitle={ICLR},
year={2024}
}

@article{liu2010learning,
  title={Learning to detect a salient object},
  author={T. Liu 
          and Z. Yuan and 
          J. Sun and 
          J. Wang 
          and N. Zheng 
          and X. Tang, Xiaoou 
          and H. Shum},
  journal={TPAMI},
  volume={33},
  number={2},
  pages={353--367},
  year={2010},
  publisher={IEEE}
}

@article{rony2023deep,
    title ={Deep Weakly-Supervised Learning Methods for Classification and Localization in Histology Images: A Survey},
    author={Rony, J. and Belharbi, S. and Dolz, J. and Ben Ayed, I. and McCaffrey, L. and Granger, E.},
    journal = "MLBI",
    volume = "2",
    year = "2023",
    pages = "96--150"
}

@inproceedings{WSOLasMIL2,
  title={Weakly supervised object localization using size estimates},
  author={Shi, Miaojing and Ferrari, Vittorio},
  booktitle={ECCV},
  pages={105--121},
  year={2016},
  organization={Springer}
}

@inproceedings{WSOLasMIL3,
  title={Multi-fold mil training for weakly supervised object localization},
  author={Gokberk Cinbis, Ramazan and Verbeek, Jakob and Schmid, Cordelia},
  booktitle={CVPR},
  year={2014}
}

@inproceedings{OldWSOL1,
  title={Localizing objects while learning their appearance},
  author={Deselaers, Thomas and Alexe, Bogdan and Ferrari, Vittorio},
  booktitle={ECCV},
  year={2010},
}

@article{murtaza2023dips,
  title={DiPS: Discriminative pseudo-label sampling with self-supervised transformers for weakly supervised object localization},
  author={S. Murtaza and 
          S. Belharbi and 
          M. Pedersoli and 
          A. Sarraf and 
          E. Granger},
  journal={Image and Vision Computing},
  volume={140},
  pages={104838},
  year={2023},
  publisher={Elsevier}
}

@InProceedings{Murtaza_2023_WACV,
  author      = {S. Murtaza and 
                  S. Belharbi and
                  M. Pedersoli and 
                  A. Sarraf and 
                  E. Granger},
  title        = {Discriminative Sampling of Proposals in Self-Supervised Transformers for Weakly Supervised Object Localization},
  booktitle    = {WACV Workshops},
  month        = {January},
  year         = {2023},
}

@article{murtaza2022constrained,
  title={Constrained Sampling for Class-Agnostic Weakly Supervised Object Localization},
  author={Murtaza, Shakeeb and Belharbi, Soufiane and Pedersoli, Marco and Sarraf, Aydin and Granger, Eric},
  journal={arXiv preprint arXiv:2209.09195},
  year={2022}
}
